\documentclass[a4paper,fleqn]{cas-sc}
\usepackage[numbers]{natbib}
\usepackage{graphicx}
\usepackage{booktabs} 
\usepackage{multirow} 
\usepackage{makecell} 

\usepackage{booktabs} 
\usepackage{amsmath}
\usepackage{amsfonts}
\usepackage{algorithm}
\usepackage{algpseudocode}
\usepackage{microtype}
\usepackage{float}

\def\tsc#1{\csdef{#1}{\textsc{\lowercase{#1}}\xspace}}
\tsc{WGM}
\tsc{QE}
\tsc{EP}
\tsc{PMS}
\tsc{BEC}
\tsc{DE}

\begin{document}
\let\WriteBookmarks\relax
\def\floatpagepagefraction{1}
\def\textpagefraction{.001}
\shorttitle{}
\shortauthors{Tingpeng Zhang et~al.}


\title [mode = title]{Mechanical in-sensor computing: a programmable meta-sensor for structural damage classification without external electronic power}  



\author[1]{Tingpeng Zhang}
\ead{tzhang876@connect.hkust-gz.edu.cn}
\credit{Conceptualization, Methodology, Software, Investigation, Formal analysis, Writing – original draft}
\affiliation[1]{organization={Intelligent Transportation Thrust, The Hong Kong University of Science and Technology (Guangzhou)},
                addressline={1 Duxue Road, Nansha District}, 
                city={Guangzhou},
                postcode={511453}, 
                state={Guangdong},
                country={China}}
                
\author[2]{Xuzhang Peng}
\ead{xpeng842@connect.hkust-gz.edu.cn}
\credit{Software, Formal analysis}
\affiliation[2]{organization={Internet of Things Thrust, The Hong Kong University of Science and Technology (Guangzhou)},
                addressline={1 Duxue Road, Nansha District}, 
                city={Guangzhou},
                postcode={511453}, 
                state={Guangdong},
                country={China}}

\author[2]{Mingyuan Zhou}
\ead{mzhou151@connect.hkust-gz.edu.cn}
\credit{Methodology, data analysis, review}

\author[2]{Guobiao Hu}
\cormark[1]
\ead{guobiaohu@hkust-gz.edu.cn}
\credit{Conceptualization, Writing - review \& editing}

\author[1,2,3]{Zhilu Lai}
\cormark[1]
\ead{zhilulai@ust.hk}
\credit{Conceptualization, Funding acquisition, Supervision, Writing – review \& editing}

\affiliation[3]{organization={Department of Civil and Environmental Engineering, The Hong Kong University of Science and Technology},
                city={Hong Kong},
                country={China}}

\cortext[cor1]{Corresponding authors. Email address: zhilulai@ust.hk (Z. Lai), guobiaohu@hkust-gz.edu.cn (G. Hu). The manuscript was submitted to an international journal for peer review.}

\begin{abstract}
Structural health monitoring (SHM) typically involves sensor deployment, data acquisition, and data interpretation, commonly implemented via a tedious wired system. The information processing in current practice majorly depends on electronic computers, albeit with universal applications, delivering challenges such as high energy consumption and low throughput due to the nature of digital units. In recent years, there has been a renaissance interest in shifting computations from electronic computing units (e.g., Graphics Processing Unit) to the use of real physical systems, a concept known as \textit{physical computation}. This approach provides the possibility of thinking out of the box for SHM, seamlessly integrating sensing and computing into a pure-physical entity, without relying on external electronic power supplies, thereby properly coping with resource-restricted scenarios. The latest advances of metamaterials (MM) hold great promise for this proactive idea. In this paper, we introduce a programmable metamaterial-based sensor (termed as \textit{MM-sensor}) for physically processing structural vibration information to perform specific SHM tasks, such as structural damage warning (binary classification) in this initiation, without the need for further information processing or resource-consuming, that is, the data collection and analysis are completed in-situ at the sensor level. We adopt the configuration of a locally resonant metamaterial plate (LRMP) to achieve the first fabrication of the MM-sensor. We take advantage of the bandgap properties of LRMP to physically differentiate the dynamic behavior of structures before and after damage. By inversely designing the geometric parameters, our current approach allows for adjustments to the bandgap features. This is particularly effective for engineering systems with a first natural frequency ranging from 9.54 Hz to 81.86 Hz; a wider range can be achieved with extended design choices. Both simulations and laboratory experiments are carried out to validate the applicability of the proposed MM-sensor, successfully achieving binary damage classification of structures in a purely physical manner and realizing the concept of mechanical in-sensor computing. 

\end{abstract}



\begin{highlights}
\item A meta-sensor is developed for in-sensor structural binary damage classification.
\item An inverse design framework is developed to programmatically calibrate the meta-sensor.
\item The function of the meta-sensor performing mechanical computing is validated by experiments.
\end{highlights}


\maketitle

\section{Introduction}

Damage detection and classification is a critical research area in structural health monitoring (SHM) for civil, aerospace, and mechanical engineering, with significant value for life safety and economic efficiency \cite{li2024mechanics,huang2024tunable,ri2024drone}. Currently, date-driven SHM \cite{rosales2017data} is a trending research effort since various sensory data (e.g., strain \cite{xie2025smart}, vibration \cite{zhou2024structural}, and temperature \cite{fan2020structural}) can be readily used to estimate the structure’s intrinsic physical attributes, such as material fatigue \cite{qu2024high}, modal properties \cite{cronin2024bridging}, and remaining service life \cite{bhattacharya2025optimal}. A typical SHM framework is generally comprised of three subsystems: sensor devices, data transmission modules, and data interpretation subsystems. Sensor devices are responsible for reliably capturing raw information, such as vibration, strain, and temperature, from the structures being monitored. The data transmission module acts as a communication hub, ensuring the flow of raw information from sensors to a central processing unit. Data interpretation is particularly important, requiring the use of advanced algorithms and models to analyze the measured data systematically. The goal is to assess the structural health status and detect potential damage. 

\begin{figure}
\centering\includegraphics[width=.99\textwidth]{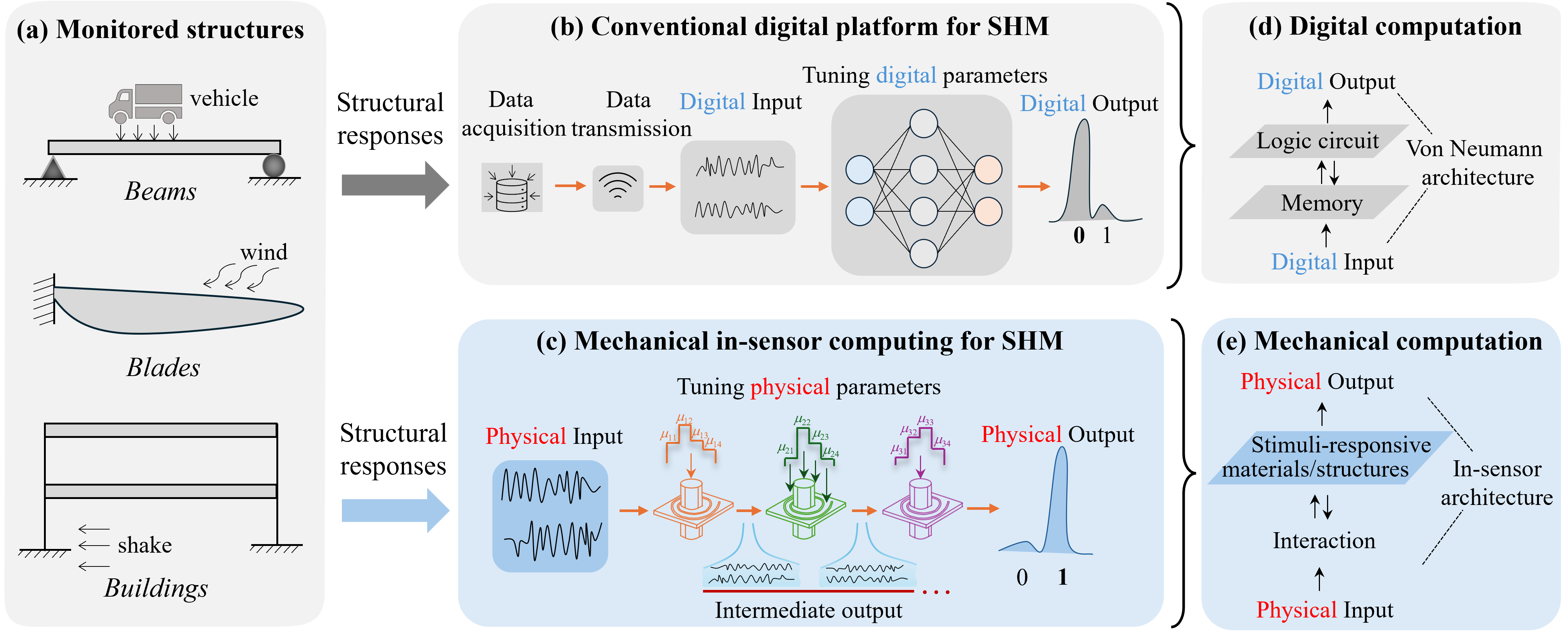}
    \caption{Comparisons between conventional digital processing-based structural health monitoring (SHM) and the proposed mechanical in-sensor computing for SHM.}
    \label{fig:frame}
\end{figure}

Recent machine learning methods have increasingly been used to automatically extract so-called "damage-sensitive features",  improving the automation and effectiveness, compared to conventional data analysis methods. 
Most recent data-driven monitoring frameworks are built on artificial neural networks -- a digital processing framework -- to distill useful patterns from massive measured data for making appropriate decisions, that is, end-to-end mapping from raw information to specific targets, as illustrated in 
Figure \ref{fig:frame}(b). However, this digital processing-based methodology comes with the following limitations: i) Training such an overparameterized neural network requires adjusting model parameters to minimize the model's outputs with desired targets. The number of parameters is severely redundant -- this high flexibility renders the models capable of learning diverse phenomena, however, making the training highly computationally expensive. Recent research \cite{malekloo2022machine} indicates that current SHM platforms based on electronic acquisition and computational units are becoming incapable of meeting the training and deployment needs of larger-scale machine learning models, to support future intelligent health monitoring systems.  ii) The current architecture of SHM relies on data acquisition, followed by data communication and analysis, ultimately leading to decision-making. This non-integrated platform may pose challenges for future real-time tasks. iii) In harsh environmental conditions \cite{lee2017toward}, such as those encountered in offshore applications, electronic devices are prone to being vulnerable.   

To address the limitations of conventional electronic/digital information processing paradigms, researchers have begun to seek alternative computing platforms with faster speed and lower power consumption, such as \textit{physical computation} \cite{shagrir2022nature}. 
These unusual physical computing systems are fundamentally different from digital electronic platforms based on the von Neumann architecture \cite{von1993first} (as illustrated in Figure \ref{fig:frame}(d)); they break the traditional hardware-software division, and perform information processing based on their physical interactions with surrounding environments (such as light pulse propagation \cite{mcmahon2023physics}, heat transfer \cite{pfeffer2022hybrid} and mechanical vibration \cite{yasuda2021mechanical}). This method directly utilizes physical phenomena to perform computational tasks faster with less energy consumption than numerical simulations or digital surrogate models. In this context, by training the physical properties (such as material characteristics) of controllable physical systems, it is possible to construct physical neural networks (PNN) that can perform tasks commonly accomplished by conventional neural networks. For example, PNNs based on electronics \cite{jeong2024cryogenic} and optics \cite{mcmahon2023physics} have advanced quickly in recent years, and can already handle certain basic classification jobs, such as classifying image dataset using ultrafast optical second-harmonic generation \cite{wright2022deep}, and completing vowel classification on spintronic nano-oscillators \cite{romera2018vowel}. 

Mechanical computation \cite{yasuda2021mechanical} (Figure \ref{fig:frame}(e)) has risen to be a moderate candidate for physical computation. The input signal directly interacts with stimuli-responsive materials or structures through physical interactions, such as mechanical vibrations, fluid flow, or heat transfer. By constructing their physical properties (such as mass distribution, stiffness, etc.), a unique processing mechanism can be formed to encode input signals. For example, the degree of deformation of a spring or the motion resistance of a link can serve as a threshold mechanism, similar to the classification logic of a simple perceptron \cite{mead1990neuromorphic}. The information transformation process from input to output is entirely embedded in the physical entity, and revealed in the physical response of the mechanical system, without relying on digital operations or logical circuits realized by conventional computational platforms. 
However, the mechanical computation has developed slowly due to the persistent lack of mature fragmented components and a reliable operational logic framework \cite{yasuda2021mechanical}. 
Most research focuses on employing mechanical processes to create logic units for binary processing, which involve switching between `off' and `on' (i.e., `0' and `1') states to represent, process, and store information. For example, by utilizing the mechanical snap-through of the unit cell composed of clamped beams \cite{song2019additively}, the transition between `0' and `1' states can be accomplished under load and constraints. These mechanical bits can also be built using the rotation of cross-sections in triangulated cylindrical origami structures \cite{treml2018origami}, as well as the "pop-up" movement of the waterbomb fold pattern's central vertices up and down. The commonality among these devices lies in their utilization of quasi-static deformations occurring between equilibrium states, which allows for the manipulation of binary information with minimal reliance on external energy. 

This binary logic functionality can also be realized by dynamic properties in mechanical systems. For example, the bistability is manifested in a micro-cantilever through the hysteretic response induced by geometric nonlinearities at large amplitudes, resulting from the forward or backward sweep of the input drive voltage \cite{venstra2010mechanical}. The emerging field of mechanical metamaterials has proposed some typical methods and construction modules to achieve the abstraction of (acoustic) mechanical computation by precisely controlling mechanical energy flow and oscillation. For instance, the developed acoustic transistors \cite{bilal2017bistable} and granular acoustic switches \cite{li2014granular} can achieve logic operations for information processing, including signal gating, switching, and cascading. Also, kirigami-based mechanical petal-shaped metamaterials \cite{wu2024mechanical} can exhibit distinct dynamic displacement states under varying mechanical forces, enabling the representation of handwritten digits `0' and `1'. The mapping based on wave physics and recurrent neural networks indicates that numerical metamaterial models can be trained to learn complex features in time-series data, completing tasks such as vowel classification \cite{hughes2019wave} and vibration recognition \cite{jiang2023metamaterial}. However, the boundary assumptions and material properties of this model are very idealistic, making it nearly impossible to realize the physical fabrication and its application in real-world scenarios.

In summary, leveraging the inherent properties of physical systems to perform computations and process information within actual physical entities is becoming a renaissance in addressing the challenges associated with conventional digital computing. The framework to successfully develop a mechanical computing system to perform specific tasks is summarized as follows: i) A critical prerequisite is the establishment of a mechanism that delivers a well-defined and stable stimuli-responsive relationship. This ensures predictable and repeatable behaviors under external stimuli, enabling the design of materials or structures with tailored dynamic properties. In this context, the burgeoning field of mechanical metamaterials offers a versatile toolset of methods to precisely manipulate mechanical energy flow, guide elastic waves, and tune the band structure \cite{yasuda2021mechanical}. ii) Solely relying on a designer’s engineering experience is often insufficient to design mechanical computational systems. To address this challenge, the development of learning rules is essential. Currently, two primary learning paradigms exist: physical learning \cite{hebb2005organization} and digital learning \cite{momeni2024training}. Physical learning requires systems' self-learning capability to autonomously and independently regulate inherent physical processes, which is currently inapplicable to mechanical computational systems. Consequently, the physical properties of these systems must be digitized, enabling iterative updates through methods analogous to the training of traditional artificial neural networks. Once the system is successfully trained, the digitized parameters can be precisely mapped to their physical counterparts, enabling the fabrication of a physical entity capable of performing specific tasks in real-world applications.

Structural health monitoring (SHM) \cite{farrar2007introduction} is a typical task that comprises sensor deployment, data acquisition, and data interpretation, commonly organized through a tedious wired system. This process can be considered as transforming structural responses (either static or dynamic) into meaningful metrics that provide insights into the health condition of structures. In this paper, we propose an \textit{unconventional} framework for structural health monitoring based on mechanical in-sensor computing (as shown in Figure \ref{fig:frame}(c)).
We explore the feasibility of using mechanical metamaterials as "sensors" (termed as \textit{MM-sensor}) to physically process structural dynamic responses and directly perform binary classification tasks in SHM. Specifically, we adopt the configuration of a locally resonant metamaterial plate (LRMP) \cite{zhang2013low} to achieve
the first implementation of the MM-sensor. It can be conveniently mounted on mechanical/engineering structures under monitoring, and its bandgap characteristics enable clear differentiation of structural dynamic behavior before and after damage. 
In our work, a numerical model of the MM-sensor is first established to validate the performance. Simulated multi-degree-of-freedom structural responses under healthy and damaged states are fed as excitations to excite the MM-sensor. The distinct responses under the two states prove the binary classification ability of the MM-sensor. We further experimented with a 3D printed prototype to confirm the validity in a laboratory environment. Notably, our proposed solution is implemented without external electronic power. The monitored structure in damaged states will deliver an excessive vibration on the MM-sensor, and the accumulated kinetic energy is converted into electrical energy via piezoelectricity, to drive any low-power consumption electrical units (such as a light) to provide an alarm of damage. To the best of our knowledge, the proposed work appears to be among the earliest efforts to explore mechanical in-sensor computing for the area of SHM. 

\section{Methodology}
In this section, we introduce a mechanical metamaterial-based methodology for binary classification of vibration signals, leveraging the unique mechanical (acoustic) properties of a locally resonant metamaterial plate (LRMP). The following subsections sequentially explain the binary mechanism and design method of the LRMP.

\subsection{ Mechanical metamaterial-based binary classifier}\label{ Mechanical metamaterial-based binary classifier}

\begin{figure}[h]
    \centering
    \includegraphics[width=.7\columnwidth]{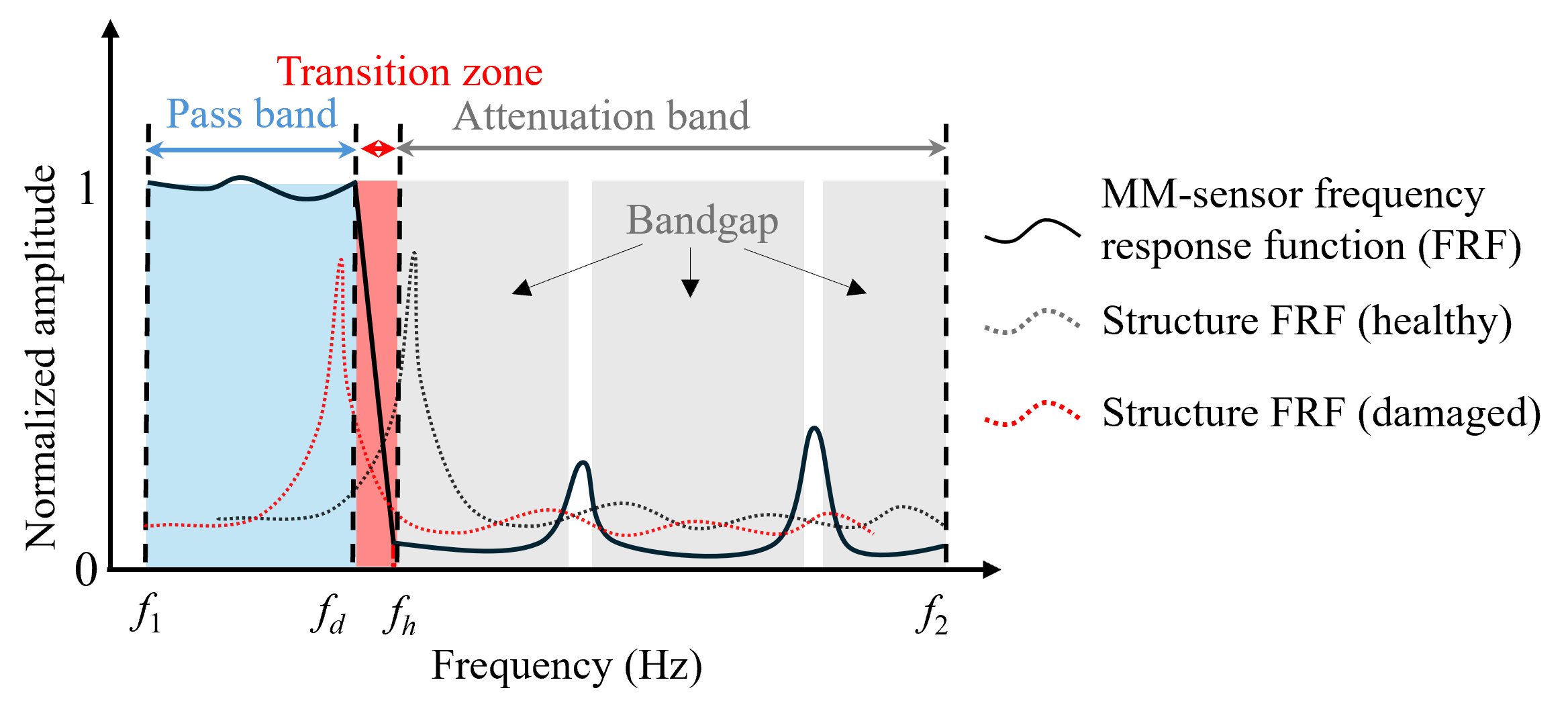}
    \caption{Illustration of the mechanism of metamaterial-based binary structural damage classifiers.}
    \label{fig:binary}
\end{figure}

The semiconductor components in electronic computing systems display `0' and ‘1' states under different voltage applications, while mechanical metamaterial-based sensors can produce binary states through distinct dynamic input-output mappings, which refers to the relationship between the input stimulus (e.g., excitation frequency, force amplitude) and the resulting output response (e.g., displacement and strain). As shown in Figure \ref{fig:binary}, a unique pass band ($f_1 \sim f_d$) - attenuation band ($f_h \sim f_2$) can be constructed within a defined frequency range of $f_1 \sim f_2$. When the mechanical vibration of the monitored structure propagates from one end of the metamaterial to the other, the excitation amplitude of elastic waves with different fundamental frequencies is amplified (within the passband) or attenuated (within the attenuation band), resulting in the vibration characteristics at the detection point forming distinct states of 0 and 1. 

As for the design principle of the propsoed metamaterials-based sensor (\textit{MM-sensor}) for structural damage classification, the frequencies of the first few modes (healthy state) that contain most of the energy of the monitored structure should be restricted within the designed attenuation band to achieve an `off' state with minimal energy response. Usually, it is pretty difficult to generate a single bandgap that is wide enough, thus multiple adjacent bandgaps may be designed to form the $f_h \sim f_2$ attenuation region. In this consideration, as long as the first natural frequency of the structure is sufficiently close to the left edge ($f_h$) of this attenuation region, any frequency shift caused by structural damage could potentially move it into the passband. This transition would facilitate a rapid release and amplification of vibrational energy, thereby serving as an effective warning evidence for damage. The red region is referred to as the transition zone ($f_d \sim f_h$ ), which reflects the rate at which the output response transitions from an attenuated state to an amplified state. This relatively narrow frequency range ensures the sensitivity of the energy states switching. A specific design is also possible to match the sensitivity needs of different monitoring scenarios.
Accordingly, the MM-sensor designed for SHM applications should meet the following criteria: 1) A tunable operating frequency below 200 Hz, aligning with the typical frequency range of most engineering structures (in contrast to conventional metamaterials, which often operate in the hundreds or thousands of Hz). 2) Flexible programmability, enabling achieving desired dynamic properties through tuning the geometric parameters of the MM-sensor.

\subsection{Inverse design of MM-sensor}\label{ Localized resonant phononic crystal plate}
\begin{figure}[h]\centering\includegraphics[width=.95\columnwidth]{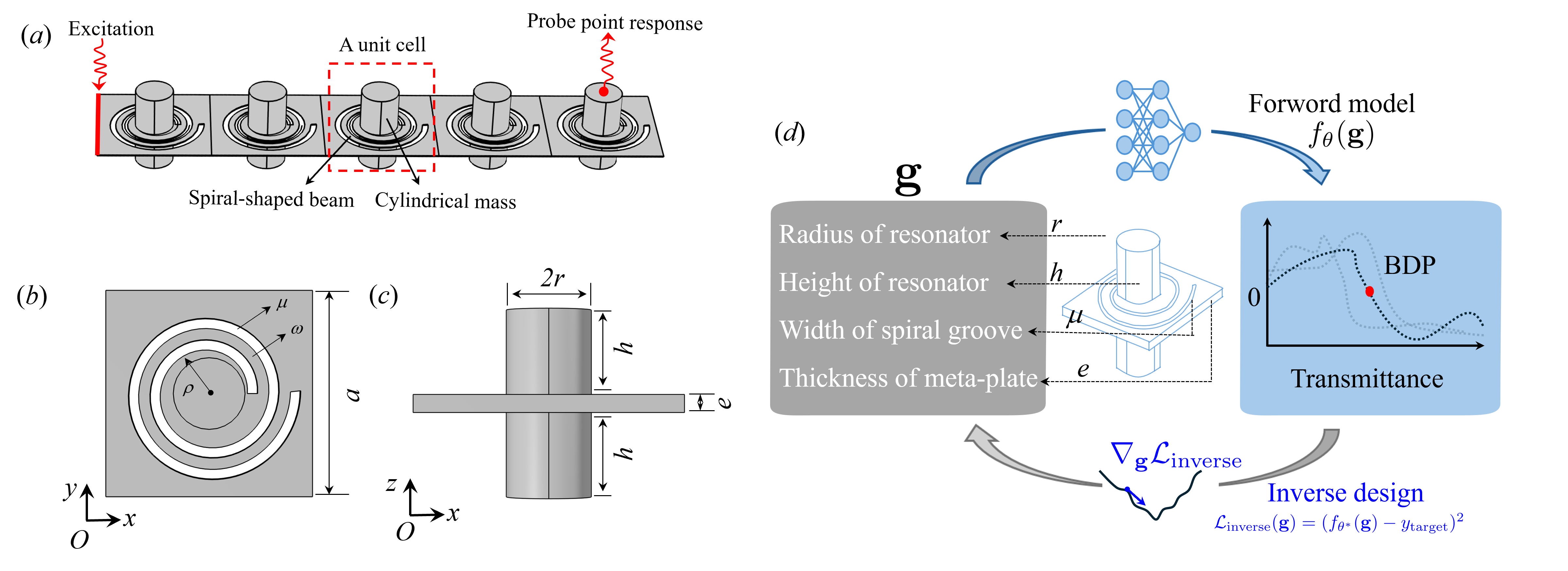}
    \caption{Design of the proposed MM-sensor: (a) 3D perspective view; (b) Top view of a unit cell; (c) Front view of a unit cell; (d) Illustration of the proposed inverse design framework ("BDP": Binary decision point).}
    \label{fig:structure}
\end{figure}


In this work, a metamaterial plate with periodic spiral resonators \cite{zhang2013low} is adopted as the basic design configuration to realize the vibration signal binary classification mechanism. As shown in Figure \ref{fig:structure}(a-c), each unit cell consists of a substrate square plate and a vertically oriented cylindrical mass, which is supported by a spiral-shaped beam directly carved from the plate. In this study, we leverage the structure of the MM-sensor's transmittance curve to realize binary classification of vibration signals. As illustrated in Figure \ref{fig:structure}(a), the transmittance $\tau(\omega)$ (unit: dB) at frequency $\omega$ is defined as the ratio of the output displacement amplitude  $W_n$ (at the probe point shown in Figure \ref{fig:structure}(a)), to the harmonic excitation with frequency $\omega$ and amplitude $W_0$ (at the left end, shown in Figure \ref{fig:structure}(a)):

\begin{equation}
    \tau(\omega) = 20 \log_{10} \left( \frac{|W_n|}{|W_0|} \right) 
\end{equation}

The transmittance curve reflects the propagation capability of elastic waves at different frequencies in the MM-sensor. Specifically, the resonance peaks in the pass band occur at frequencies where the excitation frequencies align with the natural frequencies of the MM-sensor. In contrast, within the attenuation band, most energy is absorbed by the resonators, resulting in significant vibration suppression.
More details about the formation mechanism and validation of the transmittance can be found in~\ref{app:a}. 

The characteristics of the transmittance (such as peak positions, bandgap width and positions) are mainly determined by the equivalent stiffness of the spiral-shaped beam and the equivalent mass of the cylinders. Therefore, by adjusting geometric parameters (as shown in Figure \ref{fig:structure}(b), including the resonator's height $h$, radius $r$, the thickness  $e$ of the plate, and spiral groove width $\mu$), the pass band and bandgap frequencies can be continuously calibrated (see parametric analysis in ~\ref{app:b}), thereby tailoring a transmittance curve with desired properties. Within the current range of geometric parameters ($h$ between 5–20 mm, $r$ between 1–5 mm, and $e$ between 2–5 mm, and $\mu$ between 0.5–2 mm), the left boundary frequency of the first bandgap can be tuned between 9.54 Hz and 81.86 Hz.

Traditional forward design \cite{zheng2023deep} of mechanical metamaterials relies on iteratively adjusting the unit cell configuration and performing computationally intensive simulations to evaluate whether the transmittance properties meet the desired requirement.
This repetitive design process is not well-suited for our design purpose: determining the geometric parameters that yield a target transmittance curve with specified characteristics. As illustrated in Figure\ref{fig:structure}(d), we propose an inverse design framework to address this challenge.
By extracting a scalar feature -- the frequency points at which the transmittance reaches zero in the transition region, referred to as the binary decision point (BDP). Thus, the high-dimensional transmittance data are reduced to a single representative value. This simplification avoids the complexity of an ultra-high-dimensional problem while preserving the essence of the binary classification task for damage detection by distinguishing between the attenuation band and the pass band. As illustrated in Figure 
 \ref{fig:structure}(d), 
firstly, a neural network-based forward model $f_{\theta}$ parameterized by $\theta$ takes the geometric parameters $\mathbf{g}$ as input and outputs a scalar BDP value $\hat{y}$, expressed as $\hat{y} = f_{\theta}(\mathbf{g})$.
During training, the model is optimized to minimize the mean squared error loss:
\begin{equation}
\mathcal{L}(\theta) = \frac{1}{N} \sum_{i=1}^N (y_i - \hat{y}_i)^2 + \lambda \sum \|\theta\|_2^2
\label{eq:forward_loss}
\end{equation}
where \( y_i \) is the $i$th sample of the true BDP value; \( \hat{y}_i \) is its corresponding predicted BDP value; \( N \) is the number of samples; \( \lambda = 10^{-3} \) in our study is the $\ell_2$ regularization coefficient, ensuring the trained model generalizes better to unseen data \cite{ia2016deep} (e.g., new geometric parameters $\mathbf{g}$ in the test).
After training, this trained neural network is considered as an effective forward surrogate model $f_{\theta^*}$ within a defined geometric parameter space (see the results in~\ref{app:c}), which supports a more efficient generation of a BDP value given a query of geometric parameters.

Subsequently, since neural networks are differentiable, the trained forward model $f_{\theta^*}$ is utilized for inverse design of the geometric parameters $\mathbf{g}$ of the MM-sensor.
Given the target BDP value \(y_{\text{target}} \) of a desired MM-sensor, a practical solution to design its geometric parameters is to minimize the difference between the predicted and target real BDP values:
\begin{equation}
\mathcal{L}_{\text{inverse}}(\mathbf{g}) = (f_{\theta^*}(\mathbf{g}) - y_{\text{target}})^2
\label{eq:inverse_loss}
\end{equation}
where in inverse design, $\mathbf{g}$ is randomly initialized, while the trainable parameters of the forward model $f_{\theta^*}$ are fixed (denoted by $\theta^*$). The initialized geometric parameters are optimized on the gradient of the loss $\mathcal{L}_{\text{inverse}}$ with respect to $\mathbf{g}$, i.e., $\nabla_{\mathbf{g}} \mathcal{L}_{\text{inverse}}$.
The validation results indicate that the geometric parameters can be effectively calibrated via this method.
More details and results of the design framework are presented in~\ref{app:c}.

\section{MM-sensor performance for binary damage classification}\label{result}



\begin{figure}[!htbp]
    \centering
    \includegraphics[width=.95\columnwidth]{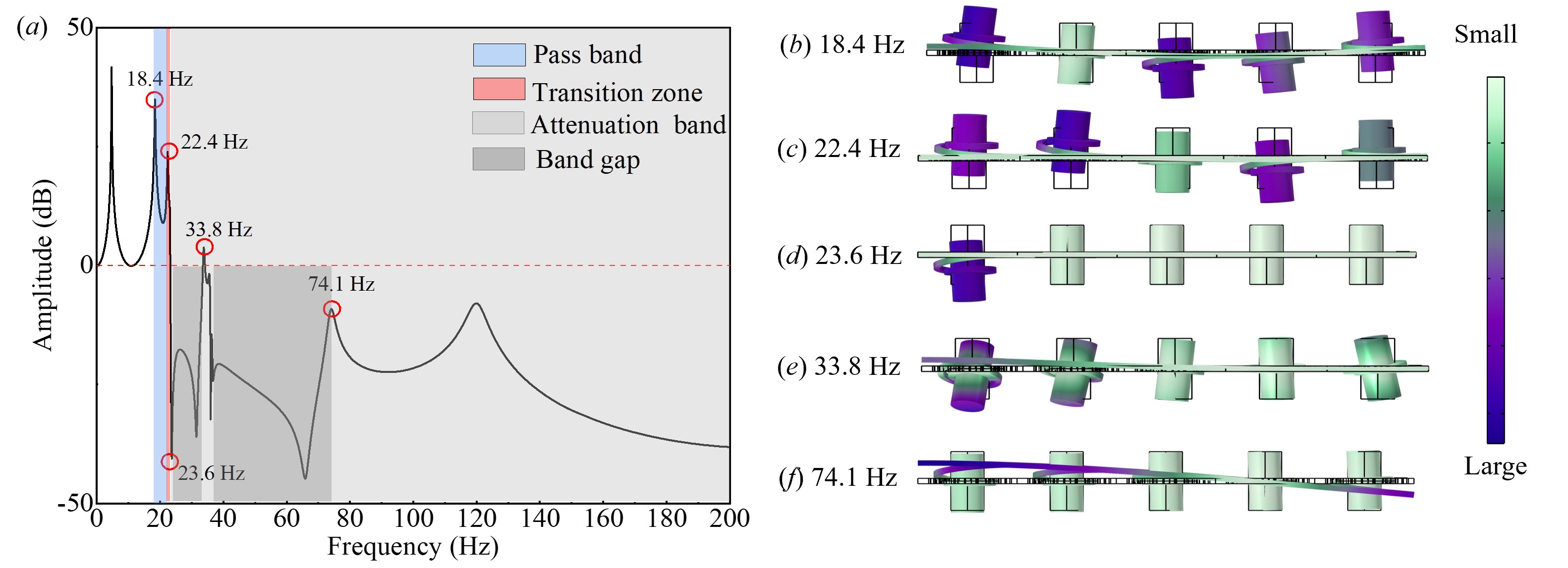}
    \caption{(a) Transmittance curve and (b-f) vibration modes of the designed MM-sensor.}
    \label{fig:desigenmeta}
\end{figure}

In this section, we take an example of monitoring a structure whose first natural frequency is 23.6 Hz. The proposed inverse design framework is applied to develop the MM-sensor with a target BDP frequency at 23.3 Hz, slightly lower than 23.6 Hz. To accommodate manufacturing constraints and
simplify the inverse design, three of the four geometric parameters are fixed ($e$ = 2 mm, $\mu$ = 1.5 mm, $r$ = 5 mm), while the resonator's height $h$ is to be optimized. The final optimized $h$ is 10.023 mm (rounded to 10 mm) to achieve the target transmittance curve with a BDP at 23.3 Hz. Using the geometric and material parameters summarized in Tables \ref{tab:table1} and \ref{tab:table2}, the transmittance of the corresponding MM-sensor obtained by finite element methods (FEM) is shown in  Figure \ref{fig:desigenmeta}(a). According to Figure \ref{fig:binary}, the pass band of this MM-sensor ranges from 18.4 Hz and 22.4 Hz, the attenuation zone spans from 23.6 Hz to 200 Hz, and the transition zone lies between 22.4 Hz and 23.6 Hz. 

\begin{table}[!htbp]
    \centering
    \begin{minipage}{0.45\textwidth}
        \centering
        \caption{Geometric parameters of a unit cell}
        \label{tab:table1}
        \begin{tabular}{lll}
            \toprule
            Definition & Notation & Value \\ 
            \midrule
            Lattice constant & $a$ & 30 mm \\ 
            Thickness of board & $e$ & 2 mm \\ 
            Width of the spiral groove & $\mu$ & 1.5 mm \\ 
            Width of spiral beam & $\omega$ & 1.51 mm \\ 
            Spiral beam's rotation angle & $\phi$ & $2\pi \sim 6\pi$ \\ 
            Radius of the cylinder & $r$ & 5 mm \\ 
            Height of the cylinder & $h$ & 10 mm \\ 
            \bottomrule
        \end{tabular}
    \end{minipage}
    \hfill 
    \begin{minipage}{0.45\textwidth}
        \centering
        \caption{Parameters of materials}
        \label{tab:table2}
        \begin{tabular}{lll}
            \toprule
            \textbf{Definition} & \textbf{Polylactic Acid} & \textbf{Steel} \\ 
            \midrule
            Density & 1520 kg/m\textsuperscript{3} & 7780 kg/m\textsuperscript{3} \\ 
            Elastic Modulus & 2.50 GPa & 210 GPa \\ 
            Poisson's Ratio & 0.360 & 0.300 \\ 
            \bottomrule
        \end{tabular}
    \end{minipage}
\end{table}

To provide further insights into the vibration pattern of the MM-sensor consisting of five unit cells, its 5 representative eigenmodes are presented in Figure  \ref{fig:desigenmeta}(b-f). 
The color rendering in these modes represents vertical displacement amplitudes, with the color bar from cooler to warmer colors indicating the vibration intensity. 
Five distinct frequencies are selected for analysis: 18.4 Hz (just before the pass band), 22.4 Hz (end of the pass band), 23.6 Hz (just before the first band gap), 33.8 Hz (midpoint between the first and second band gaps) and 74.1 Hz (end of the second band gap). 
Figures \ref{fig:desigenmeta}(b) and \ref{fig:desigenmeta}(c) show that the wave propagates evidently from one end to the other end of the MM-sensor, when the excitation frequency lies within the pass band (18.4 Hz - 22.4 Hz). 

When the excitation frequency falls within the band gaps ([23.6, 33.4] Hz and [34, 71.9] Hz), as illustrated in Figure \ref{fig:desigenmeta}(d), a distinct pattern is observed: the displacement on the right-hand side is minimal, and only the first few resonators (cylindrical masses) exhibit noticeable vibrations, while the others remain nearly stationary.
This result indicates that the vibration in the attenuation band is effectively suppressed. 
For the mode in Figure \ref{fig:desigenmeta}(f), the free end of the plate exhibits vibrations, while the cylindrical mass undergoes minimal motion. This is due to the cancellation caused by the opposite motion phases of the excited end and free-end movements. 
This also explains why the transmittance falls below 0 dB at 74.1 Hz. More information about the mechanism and the validation of this transmittance can be found in ~\ref{app:a}. 

In the following sections, a series of numerical case studies are conducted to thoroughly assess the designed MM-sensor's capability of indicating structural health conditions.


\subsection{Validation on synthesized structure response}

This section is to preliminarily validate the effectiveness of the proposed MM-sensor for structural damage binary classification. The response of the monitored structure is first emulated using a superposition of multi-frequency harmonic waves, which is assumed to reflect the vibration patterns of various mechanical structures, such as beams and slabs. The simulated signal is a superposition of five fundamental sine waves without added noise, and is expressed as:
\begin{equation}
\text{Disp}(t) = A_1 \sin(2\pi f_1 t) + A_2 \sin(2\pi f_2 t) + A_3 \sin(2\pi f_3 t) + A_4 \sin(2\pi f_4 t) + A_5 \sin(2\pi f_5 t)
\end{equation}
where \( \text{Disp}(t) \) denotes the simulated signal emulating structural displacements; \( A_1 \) to \( A_5 \) represent the amplitudes of the first to fifth modal orders, respectively, indicating the intensity of each mode; \( f_1 \) to \( f_5 \) denote the corresponding modal frequencies (in Hz).


\begin{table}[!htbp]
\centering
\caption{Three cases emulating structural responses (healthy and damaged)}
\label{tab:multi_frequency_harmonic_waves}
\begin{tabular}{@{}llccccc@{}}
\toprule
 \textbf{Case} & & \textbf{Mode 1} & \textbf{Mode 2} & \textbf{Mode 3} & \textbf{Mode 4} & \textbf{Mode 5} \\
\midrule
  & Amplitude (mm) & $1.10$ & $0.13$ & $0.05$ & $0.02$ & $0.01$ \\
Case 1 & Natural Frequencies (Hz) -- Healthy & $23.71$ & $45.16$ & $68.97$ & $91.33$ & $115.16$ \\
 & Natural Frequencies (Hz) -- Damaged & $22.41$ & $44.82$ & $67.27$ & $89.69$ & $112.09$ \\
\midrule
  & Amplitude (mm) & $1.16$ & $1.04$ & $0.29$ & $0.03$ & $0.02$ \\
Case 2 & Natural Frequencies (Hz) -- Healthy & $23.65$ & $25.34$ & $50.77$ & $91.69$ & $135.49$ \\
 & Natural Frequencies (Hz) -- Damaged & $22.46$ & $24.63$ & $49.28$ & $90.60$ & $133.11$ \\
\midrule
  & Amplitude (mm) & $0.12$ & $0.26$ & $0.53$ & $1.03$ & $0.15$ \\
Case 3 & Natural Frequencies (Hz) -- Healthy & $23.74$ & $68.39$ & $82.50$ & $91.87$ & $154.61$ \\
  & Natural Frequencies (Hz) -- Damaged & $22.48$ & $67.99$ & $80.12$ & $90.33$ & $150.66$ \\
\bottomrule
\end{tabular}
\end{table}

\begin{figure*}
    \centering
    \includegraphics[width=.95\textwidth]{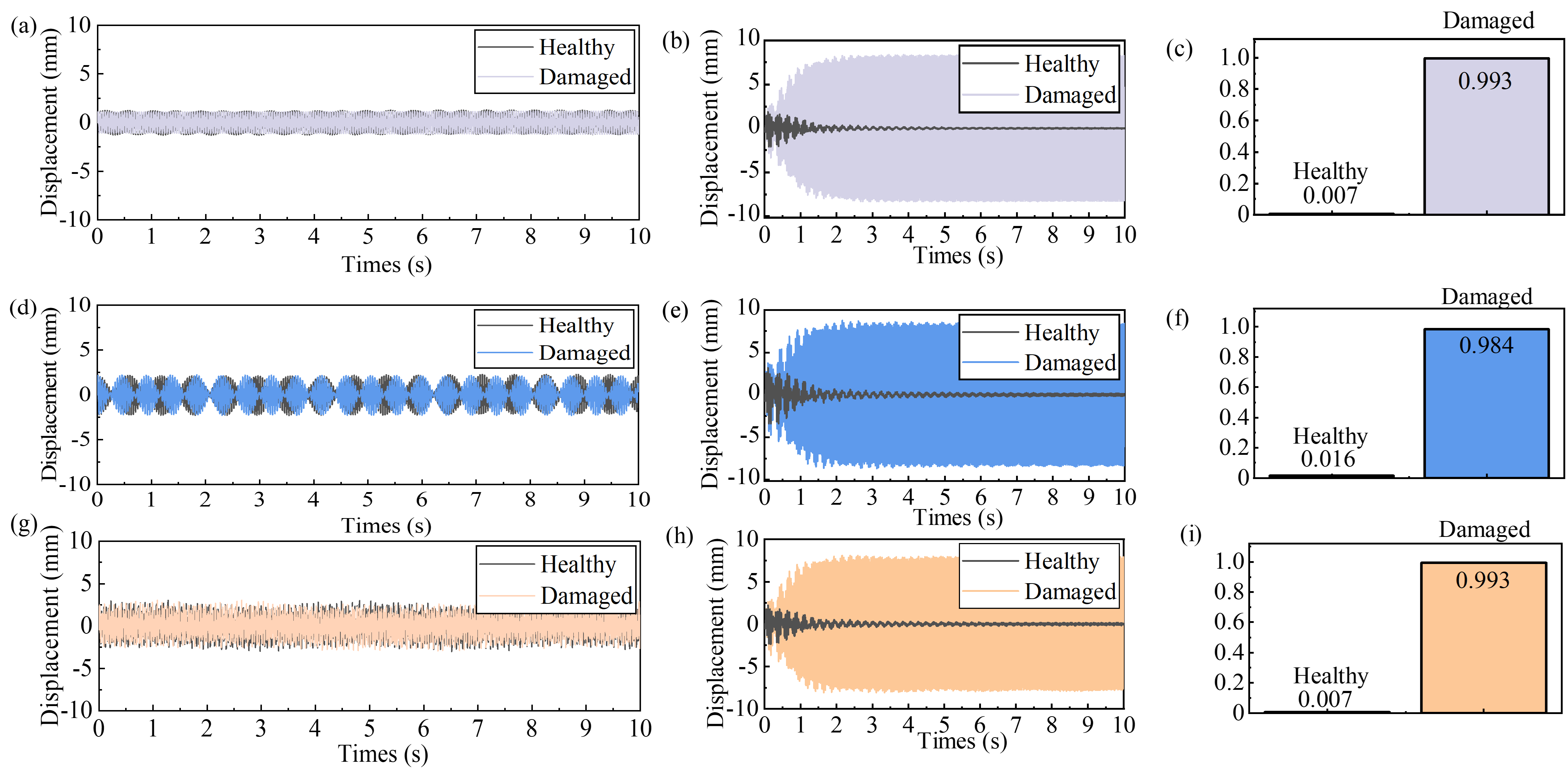}
    \caption{The structural response and the MM-sensor response at the probe point in healthy and damaged states: (a,d,g) Structural responses for Case 1, 2 and 3, respectively; (b, e, h) MM-sensor responses for Case 1, 2, and 3, respectively; (c, f, i) Classification metrics for Case 1, 2, and 3, respectively.}
    \label{fig:mutisin}
\end{figure*}

The parameters corresponding to three cases are listed in Table \ref{tab:multi_frequency_harmonic_waves}. For example, in Case 1, the response of the structure in a healthy state is mainly composed of $f_1 = 23.71$ Hz, $f_2 = 45.16$ Hz, $f_3 = 68.97$ Hz, $f_4 = 91.33$ Hz, and $f_5 = 115.16$ Hz, with corresponding amplitudes of $A_1 = 1.10$ mm, $A_2 = 0.13$ mm, $A_3 = 0.05$ mm, $A_4 = 0.02$ mm, and $A_5 = 0.01$ mm. Under the condition that the amplitude remains unchanged, each fundamental frequency is reduced by 3\% to simulate structural damage. These two types of signals are shown in Figure \ref{fig:mutisin}(a) as an example. They are used as excitation input from the fixed end to the MM-sensor, and the displacement response at the probe point is extracted as the output of the sensor. As illustrated in Figure \ref{fig:mutisin}(b), it is evident that when the structure is undamaged, the steady-state displacement amplitude at 10 seconds of the measurement point is 0.046 mm (i.e., approximately zero). In contrast, when structural damage occurs, the vibration amplitude at the probe point is significantly amplified to 16.420 mm. We assign the two displacement vibration amplitudes into a vector, and normalize the vector by the
sum of the values. Then it is served as the MM sensor’s predicted metric for the structural damage classification. It is \([0.007, 0.993]\) as visualized in Figure \ref{fig:mutisin}(c), which can be closely interpreted as the states of 0 and 1.

Next, we evaluated the performance of the MM-sensor under the other two excitations that have distinct characteristics, shown in Figure \ref{fig:mutisin}(d) (Case 2) and (g) (Case 3). The excitation in Case 2 is primarily dominated by two closely spaced modes in the low-frequency range (23.65 Hz and 25.34 Hz), which may lead to misleading results for the MM-sensor to perform signal classification. The structural response of Case 3 exhibits significant high-frequency noise (91.87 Hz with amplitude 1.03 mm), representing the types of noise encountered in practical applications due to factors such as installation looseness and environmental conditions. The results are shown in Figures \ref{fig:mutisin}(e) and (h). Clearly, for these two cases, the MM-sensor still achieves satisfactory classification performance by attenuating the structural response under healthy conditions and amplifying it under damaged conditions. 
The predicted metric of healthy and damaged states for Case 2 (Figure \ref{fig:mutisin}(f)) is $[0.016, 0.984]$, and is $[0.007, 0.993]$ for Case 3 (Figure \ref{fig:mutisin}(i)). The results indicate that even when the excitation sources exhibit distinct vibration patterns, the MM-sensor remains capable of detecting subtle frequency variations, capturing the distinct vibration amplitude responses associated with healthy and damaged conditions.

\begin{figure}
    \centering
    \includegraphics[width=.99\columnwidth]{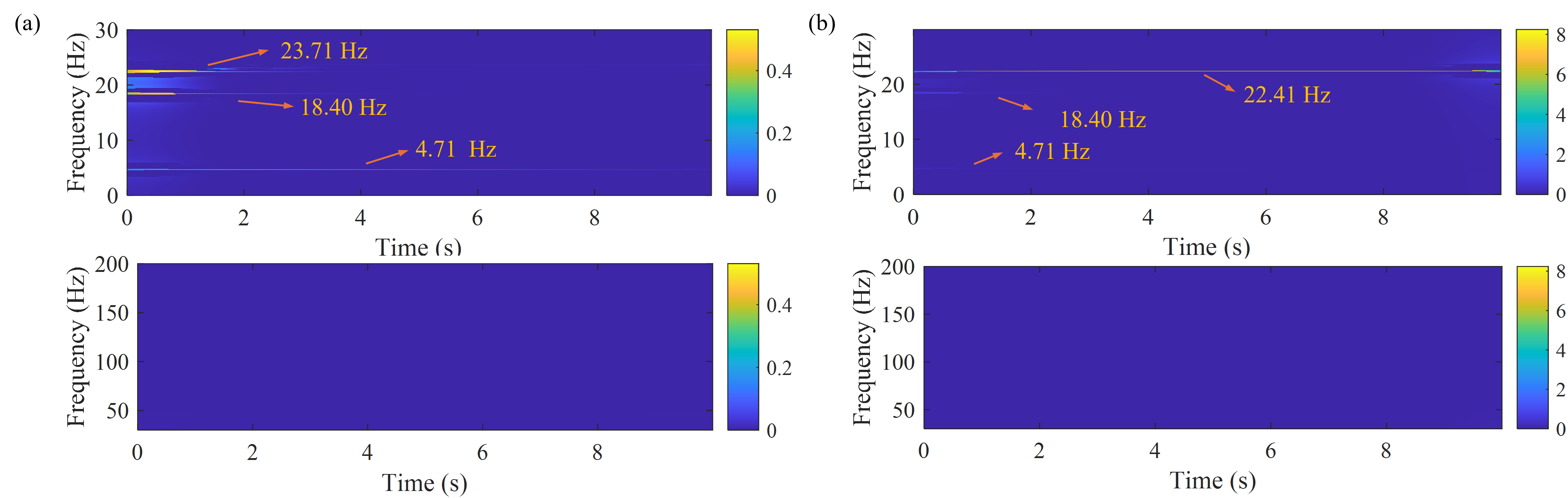}
    \caption{Time-frequency analysis of the response of the MM-sensor in Case 1: (a) Structure in healthy state; (b) Structure in damaged state.}
    \label{fig:time-fre2}
\end{figure}

To better understand the working mechanism of the MM-sensor, a time-frequency plot of the response (in Case 1) is shown in Figure \ref{fig:time-fre2}. To have better visualization, 0-30 Hz and 30-200 Hz are plotted separately in two subplots. It is clear that when the structure is undamaged, the excitation fundamental frequency (23.71 Hz) is rapidly dissipated to nearly zero within 3 seconds. Conversely, the energy is more quickly concentrated and amplified within 1 second after frequency shifts (22.41 Hz) caused by the structural damage. A similar frequency sensitivity (i.e., frequency filtering characteristics) has also been observed in biological neural systems \cite{hutcheon2000resonance}. It is noteworthy that both time-frequency diagrams deliver two other frequencies, 4.71 Hz and 18.4 Hz, which correspond to the first and second natural frequencies of the MM-sensor, respectively. Regardless of whether the monitored structure is healthy or damaged, these two modal frequencies do not interfere with the target frequency being processed by the MM-sensor, as they both quickly decay to minimal energy values within a short time. The second to fifth order modes of the emulated structure in the frequency range of 30 Hz to 200 Hz are invisible, implying that their energy is extremely low. This aligns with our expectations, which only focus on the magnitude of the structural fundamental frequency. In other words, regardless of whether the structure is healthy or damaged, other order frequencies within 200 Hz of the structure do not affect the MM-sensor's frequency filtering performance. 


\subsection{Validation on a 6-degree-of-freedom (DOF) structural system}

\begin{figure*}
    \centering
    \includegraphics[width=.99\textwidth]{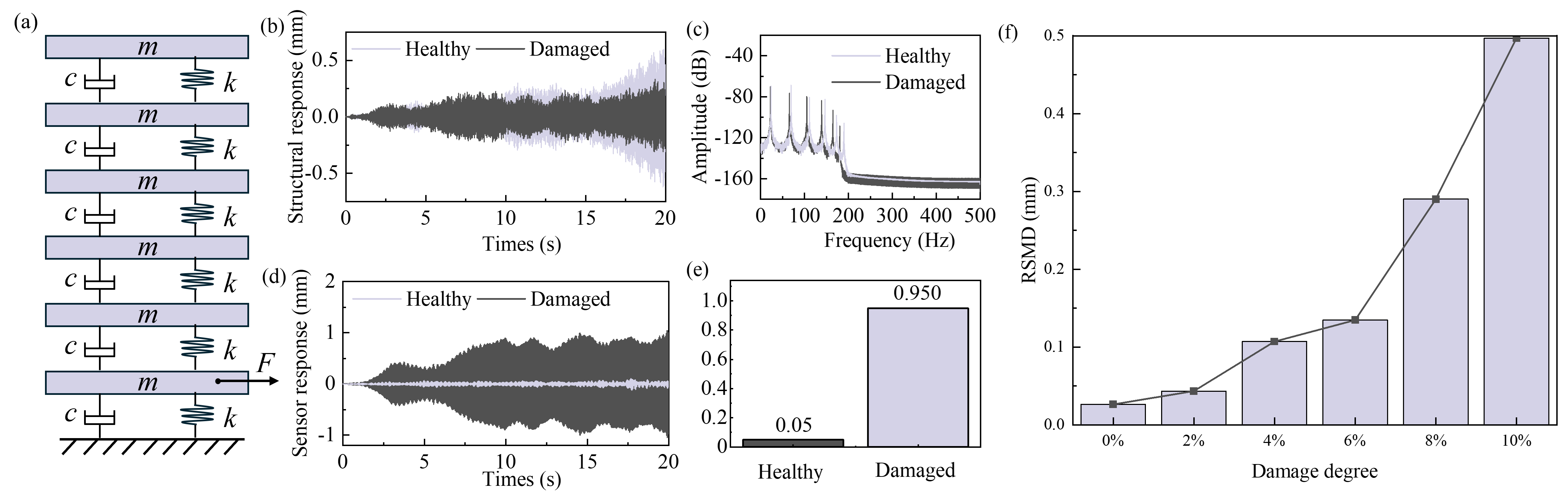}
    \caption{Case study on a 6-degree-of-freedom spring-mass-damper structure: (a) Schematic illustration; (b) Structural response in healthy and damaged states;(c) Power spectrum density of structural responses in healthy and damaged states; (d) MM-sensor probe response in healthy and damaged states; (e) classification metric; (f) Sensitivity curve.}
    \label{fig:randomexcitation}
\end{figure*}
Subsequently, to demonstrate the strong applicability of the proposed MM-sensor, we examined its performance under more complex and realistic structural responses. Specifically, we presented a numerical example of a 6-DOF structural system, as shown in Figure \ref{fig:randomexcitation}(a). This system represents a typical spring-mass model commonly used to simulate the dynamic behavior of generic engineering structures under random excitation. Each mass block has the same mass \( m_i = 50 \, \text{kg} \) (for \( i = 1, 2, \ldots, 6 \)), with an initial spring stiffness of \( k_i = 18900 \, \text{kN/m} \) (for \( i = 1, 2, \ldots, 6 \)), and a damping coefficient of \( c_i = 10 \, \text{N} \cdot \text{s/m} \) (for \( i = 1, 2, \ldots, 6 \)). The external excitation is Gaussian white noise with an amplitude of 1000 N, applied to the first mass block. The "ode45" numerical solver in Matlab \cite{senan2007brief} was used to simulate the structure's response, considering strucutral damage where the spring stiffness drops from \( 18900 \, \text{kN/m} \) to \( 17000 \, \text{kN/m} \) (approximately a 10\% reduction). As shown in Figure \ref{fig:randomexcitation}(c), the first six frequencies of this structure are all within the normal operating frequency range of the proposed sensor, which is 0-200 Hz. Due to the reduction in stiffness, the first-order frequency of the structure decreased from 23.6 Hz to 22.4 Hz. However, the displacement response of the structure (represented by the topmost mass block) exhibited no evident changes before and after the damage (Figure \ref{fig:randomexcitation}(b)). The MM-sensor is installed on the topmost mass block to monitor the structural health state, and the displacement responses recorded at the probe points are presented in Figure \ref{fig:randomexcitation}(d). It is observed that the proposed MM-sensor successfully demonstrates both attenuation and amplification states in response to the structure's healthy and damaged states. 

Unlike the synthetic multi-harmonic vibration, the random excitation signal introduces fluctuations to the amplitude of the responses, making it challenging to extract a stabilized one. In this regard, we recommend using the root mean square displacement (RMSD) of the probe point over a specific time duration (20 seconds in this example)  as the evaluation metric, which can be expressed as:
\begin{equation}
\text{RMSD} = \sqrt{\frac{1}{N} \sum_{k=1}^{N} x_k^2}
\label{eq:RMSD}
\end{equation}
where $N$  is the number of samples in the segment; \( x_k \) represents the individual displacement value in the segment.

As shown in Figure \ref{fig:randomexcitation}(d), it is computed that the RMSD at the probe point is 0.026 mm when the monitored structure is undamaged. It sharply increases to 0.497 mm — an increase of one order of magnitude — when the structure experiences a 10\% reduction in stiffness, making such a significant energy surge easily detectable. Furthermore, we considered four milder cases, specifically with stiffness reduction of 8\%, 6\%, 4\%, and 2\%. The RMSD values of the probe points under these conditions are summarized in Figure \ref{fig:randomexcitation}(f), forming a sensitivity curve of RMSD changes to the damage degree of the structure. The graph primarily shows that after around 4\% damage to the structure, the MM-sensor exhibits a more evident early warning characteristic, namely a large rise in vibration energy. Additionally, this curve also demonstrates the sensor's high sensitivity to frequency shifts caused by slight stiffness loss (as low as 2\%), which may provide insights for future multi-classification problems, aiming to further distinguish between different degrees of structural damage.

\begin{figure}
    \centering
    \includegraphics[width=.89\columnwidth]{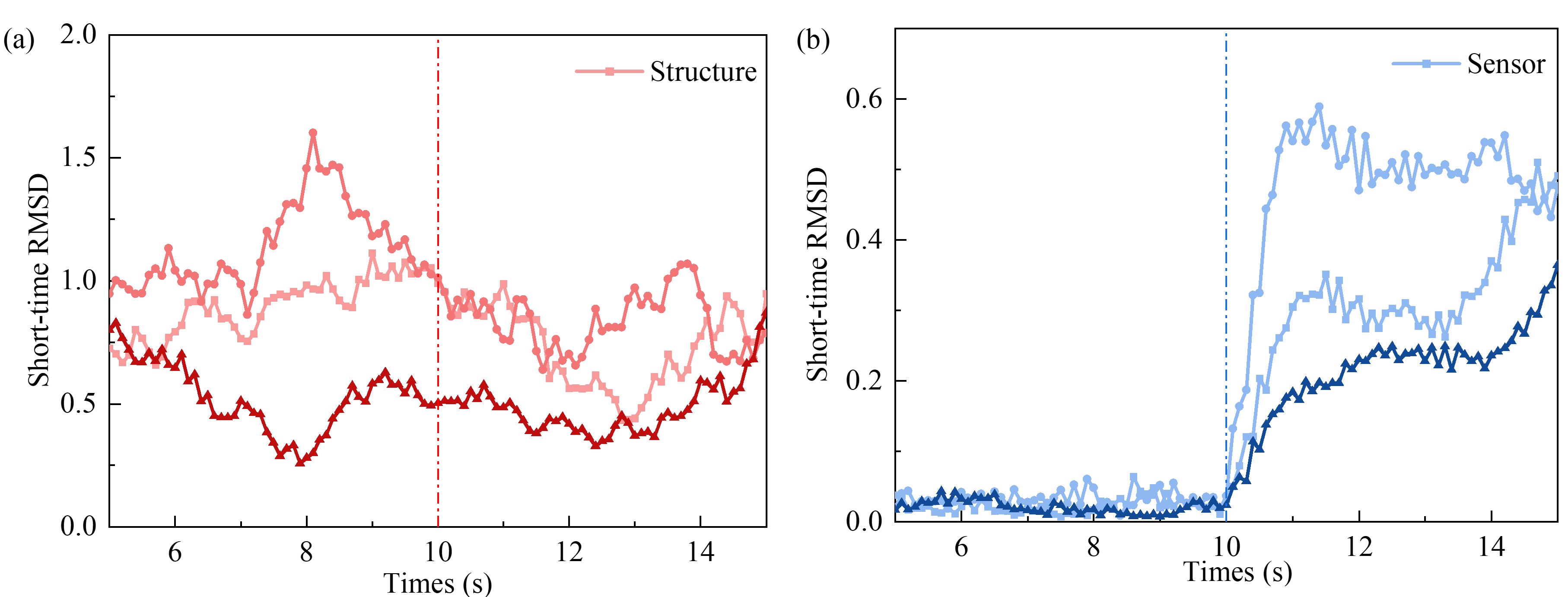}
    \caption{Short-time RMSD during sudden stiffness damage at the 10th second (three repeated experiments): (a) Structure; (b) MM-sensor.}
    \label{fig:EI_dam}
\end{figure}

As shown in Figure \ref{fig:EI_dam}, if one evaluates the RMSD every 0.1 seconds, a short-time RMSD curve for the structure and MM-sensor can be obtained (the 3 curves representing 3 repeated experiments with random excitations).  When the structure suffers 10\% stiffness damage at the 10th second, the short-time RMSD of the structural response shows no evident variations, while the proposed MM-sensor exhibits a notable energy step at the 10th second. This result is comparable to developed algorithms based on electronic processing units \cite{lai2016moving}, but the proposed MM-sensor is purely mechanical, without the need for extra data collection and transmission procedures.

\subsection{Alarm systems of damage states}\label{pzt}
\begin{figure*}
    \centering
    \includegraphics[width=.99\textwidth]{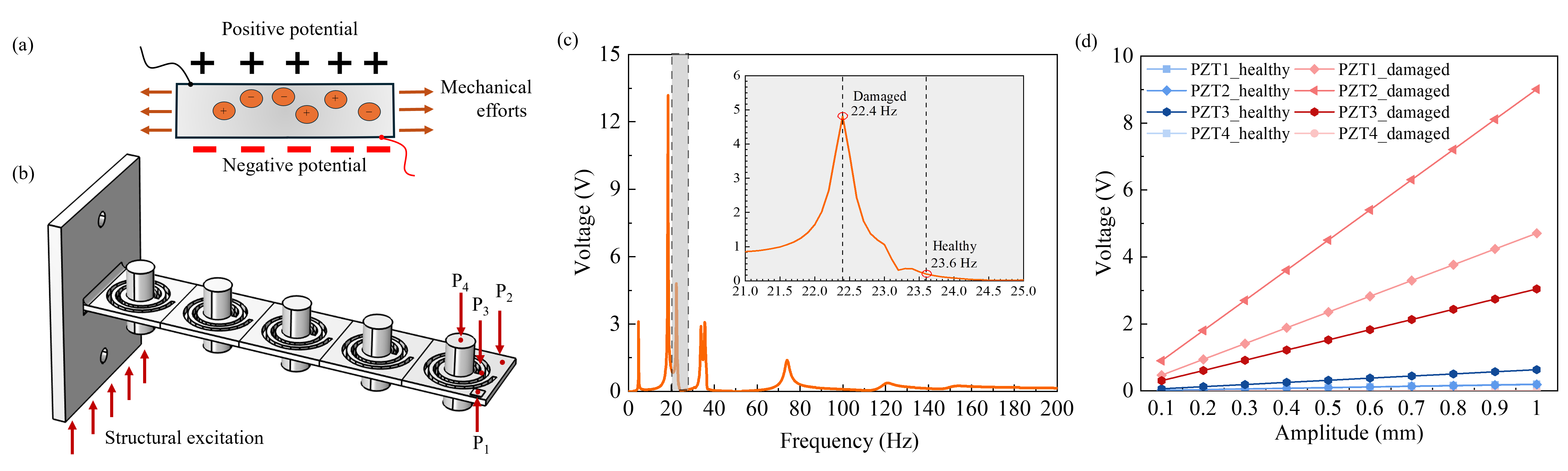}
    \caption{Piezoelectric energy harvesting for self-powered sensing: (a) \( d_{31} \) piezoelectric modes for piezoelectric materials; (b) Schematic diagram of piezoelectric element installation; (c) Voltage curve during frequency sweep experiment for location $P_2$; (d) Generated voltages under different amplitudes.}
    \label{fig:pzt12}
\end{figure*}
To acquire the classification metric, the above case requires real-time monitoring of the MM-sensor's response at the probe point. This is difficult to implement in practice. Therefore, it's necessary to develop a wireless alarm system on the MM-sensor. The piezoelectric energy harvesting is an effective method, which can convert kinetic energy, such as vibrations or impacts, into electrical energy. This method enables the alarm system to function without depending on external power sources. Specifically, the piezoelectric effect is described as the asymmetric displacement of charges or ions in piezoelectric materials (such as quartz and barium titanate), when subjected to mechanical strain, as illustrated in Figure \ref{fig:pzt12}(a). The piezoelectric effect is categorized into the direct effect and inverse effect. The direct effect is typically used in sensors and energy harvesters, while the inverse effect is applied in actuators. In this study, the direct piezoelectric effect described by the following equation is utilized:
\begin{equation}
D_k = d_{kij} T_{ij}
\end{equation}
where \( D_k \) represents the electric displacement tensor, measured in the unit of \( C\cdot m^{-2} \). The term \( d_{kij} \) denotes the piezoelectric charge coefficient tensor of the material, expressed in \( C\cdot N^{-1} \), while \( T_{ij} \) refers to the stress tensor, which is measured in \( N\cdot m^{-2} \). This work selects the most widely used piezoelectric ceramic materials due to their cost-effectiveness, robustness, and high piezoelectric performance. By gluing piezoelectric elements to the MM-sensor, a \( d_{31} \) piezoelectric mode, also known as the horizontal mode, is formed.

We conducted a simple verification to determine whether the vibration-induced strain could be converted into an AC voltage through a simple piezoelectric circuit, in order to trigger an alarm. We first conducted a sweep-frequency (0 to 200 Hz) analysis with an amplitude of 1 mm. We extracted the voltage values from the piezoelectric element (dimensions: 4 mm $\times$ 4 mm $\times$ 0.2 mm) installed at position 1 ($P_1$), as shown in Figure \ref{fig:pzt12}(b).  Figure \ref{fig:pzt12}(c) shows that the generated piezoelectric voltage exhibits significant fluctuations, delivering a pattern closely consistent with the transmittance curve, especially at the peak and trough positions. In the highlighted frequency range of 22.4 - 23.6 Hz, the voltage is minimal and insufficient for driving the warning circuit when the structure is undamaged with a frequency of 23.6 Hz. When the structure is damaged with a frequency of 22.4 Hz, the excessive vibrations at the end of the MM-sensor are capable of driving the circuit voltage to reach 4.76 V, which is sufficient to drive a custom low-power sensing circuit to transmit simple signals to indicate an alarm.

Figure \ref{fig:pzt12}(d) shows the voltage amplitude curves generated by piezoelectric elements of the same size attached at four different locations ($P_1$ - $P_4$), when subjected to structural excitations of different amplitudes (\(0.1 \, \text{mm} - 1 \, \text{mm}\)). The markers of various colors and shapes are used in the figure to differentiate between conditions, with the blue markers indicating the healthy state and the red markers indicating the damaged state. It is observed that the curve values corresponding to the damaged state are significantly higher than those under the healthy states, especially at higher excitation amplitudes. This indicates that the damaged state of the structure can be distinctly differentiated by the voltage values generated by the piezoelectric elements. As the excitation amplitude of the structure increases, the voltage values under the damaged state rise significantly, showing a clear linear growth trend at all the four positions, while the values under the healthy state change little. 
The piezoelectric element at $P_4$ (i.e., at the top surface of the stainless steel, whose strains are negligible) is unlikely to generate a significant voltage value even in the case of structural damage, making it not an ideal candidate for installing piezoelectric elements. In contrast, the piezoelectric elements at $P_1$ and $P_2$ (on the plate) can at least generate a voltage value (\(0.3 \,\text{V} - 9 \,\text{V}\)) in the event of structural damage, showing capability to drive an alarm circuit. 

\subsection{Experimental validation}\label{exp}

This section further presents experiments to assess the applicability of the proposed MM-sensor across various scenarios, with the goal of guiding future real-world applications.
\subsubsection{Experimental setup}

\begin{figure}[h]
    \centering
    \includegraphics[width=.5\columnwidth]{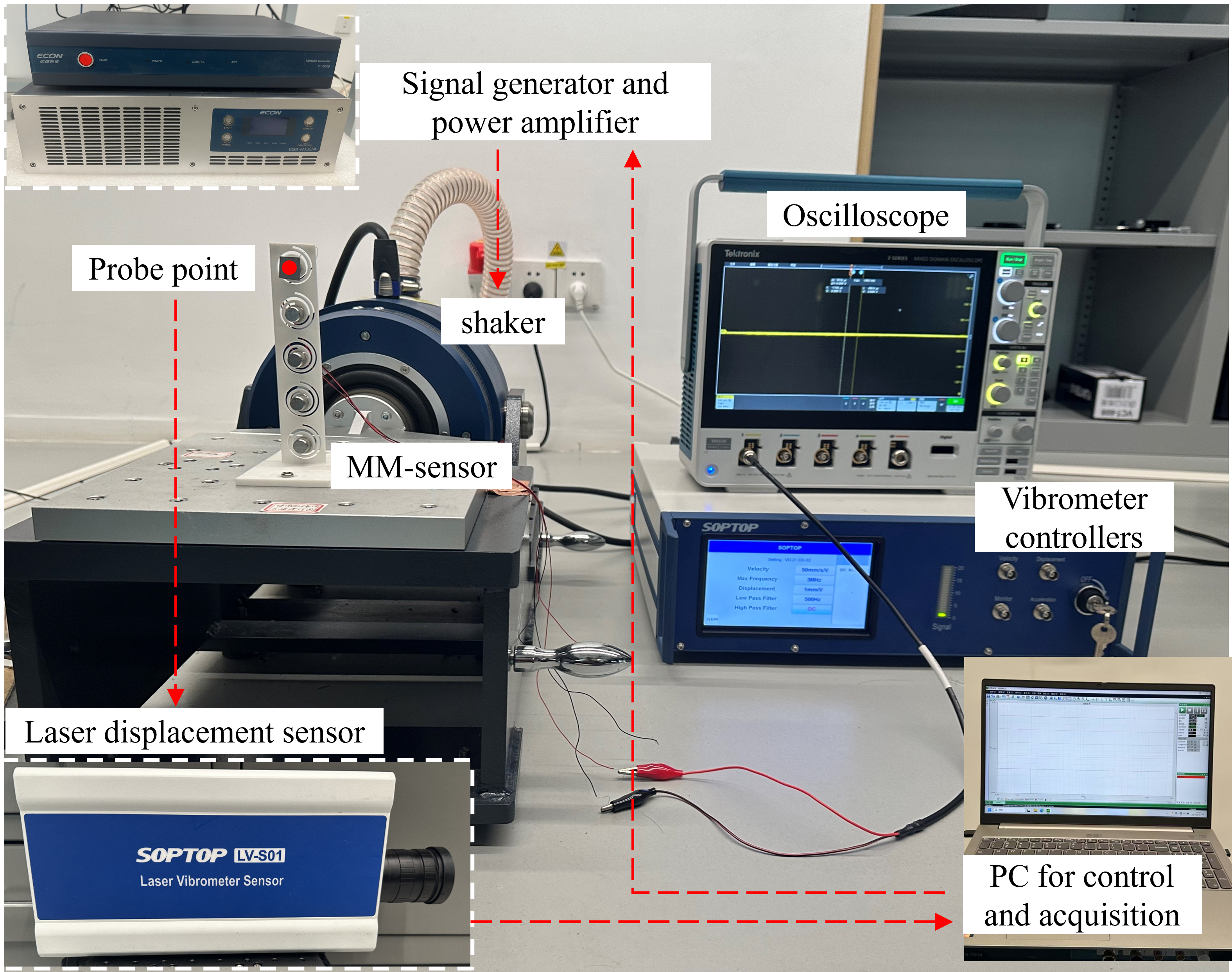}
    \caption{Overview of the experimental setup.}
    \label{fig:expset}
\end{figure}

To experimentally validate the proposed MM-sensor, a 3D-printed metamaterial plate with five resonant units was manufactured. The geometric configuration is identical to the numerical model. Figure \ref{fig:expset} shows the experimental setup. A vibration test configuration file is created to generate user-defined vibration signals. This signal is then amplified by a power amplifier, subsequently applied to a shaker to generate the corresponding excitation to the mounted MM-sensor. The shaker table can provide real-time feedback on its actual vibration conditions, facilitating program control adjustments. The center of the top cylindrical surface is the probe point, measured by a laser displacement sensor.

\begin{figure}[h]
    \centering
    \includegraphics[width=.89\columnwidth]{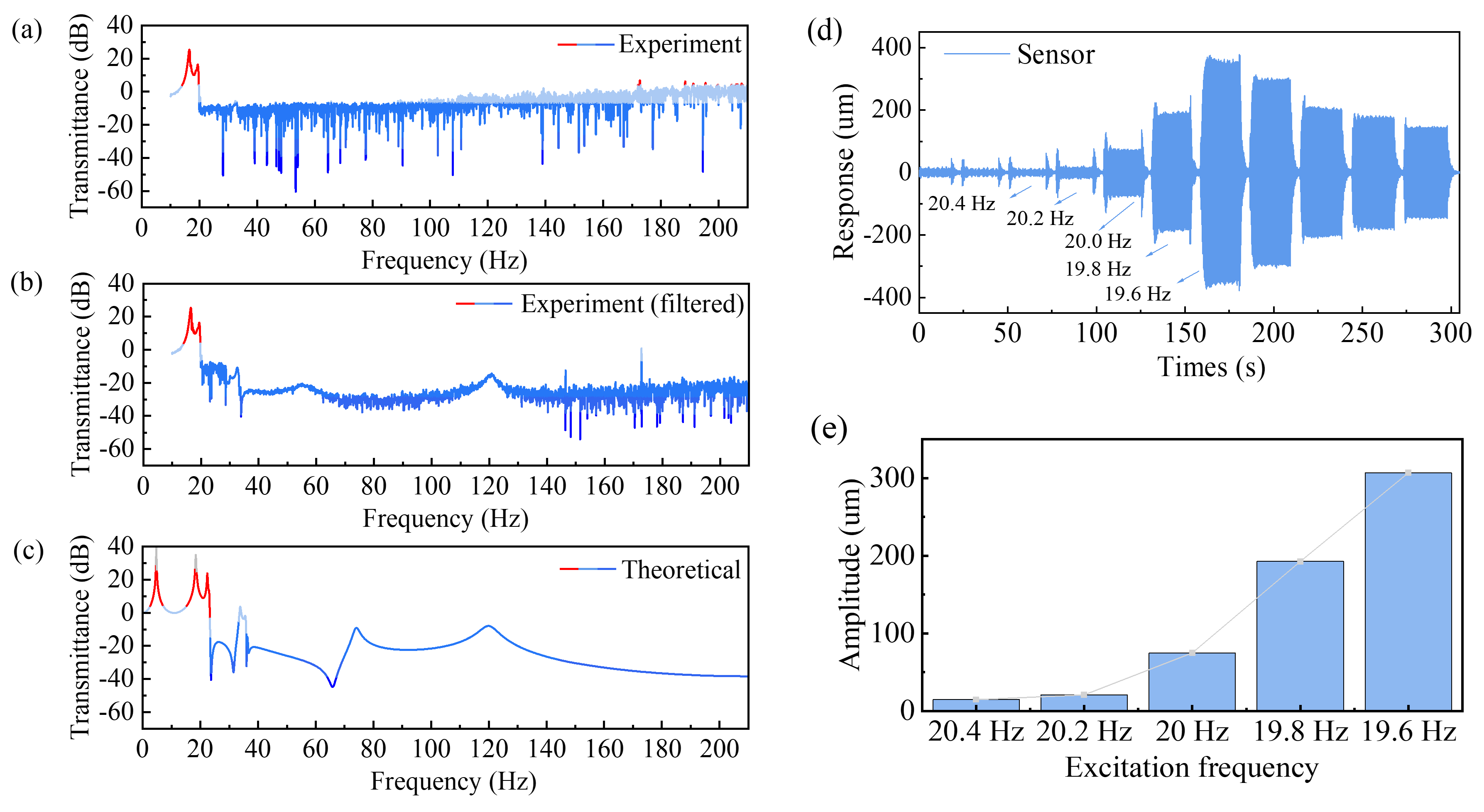}
    \caption{Experimental validation: (a) Raw transmittance obtained in the experiment; (b) Filtered transmittance; (c) Theoretical transmittance obtained by a corresponding finite element model; (d) Frequency sweep experiment (sampling every 0.2 Hz for 20 seconds); (e) The corresponding vibration amplitudes for figure
(d).}
    \label{fig:exp1}
\end{figure}

Firstly, the transmittance curve obtained by a sweep frequency (10 - 200 Hz) experiment is shown in Figure \ref{fig:exp1}(a). It is clear that it differs from the one obtained by a corresponding finite element model (in Figure \ref{fig:exp1}(c)). No distinctive peaks and valleys are observed beyond 20 Hz. This phenomenon occurs because the damping ratio at high frequencies in the experiment is greater than that at low frequencies, resulting in a significant attenuation of the vibration response. Consequently, the signal-to-noise ratio (SNR) decreases, leading to the signal being overwhelmed by noise.
Thus, a band-pass filter (with a frequency range of 20 Hz to 250 Hz)) was applied to this part of the data, and the filtered transimittance curve is shown in Figure \ref{fig:exp1}(b), which is more consistent with the theoretical one. The filtered experimental results clearly deliver the presence of pass band (16.5 Hz - 19.6 Hz), transition zone (19.6 - 20.4 Hz), and attenuation band (20.4 Hz - 200 Hz). These values are slightly lower than the corresponding theoretical results. Several factors may contribute to this discrepancy.  The manufacturing imperfections in the thickness of the plate and the width of the spiral beam are the primary contributors. Also, achieving boundary conditions that fully match the theoretical model in a laboratory environment is challenging. It is worth noting that the experimental transmittance value is slightly lower than the theoretical prediction in the low-frequency region, and slightly higher than the theoretical value in the high-frequency region. This difference is mainly attributed to the excessive damping in the finite element model, coupled with the significant damping effect in the high-frequency domain \cite{zhao20223d}.



Albeit the discrepancies, this experimental MM-sensor still delivers a clear pass band and attenuation band that can be effectively used for binary structural damage classification. We experimentally identify that the transition zone is from 20.4 Hz to 19.6 Hz (from high to low). 
As evidence, we plot the response amplitude at the probe point and its corresponding excitation frequency in Figures \ref{fig:exp1}(d) and (e). The figures show that the amplitude significantly increases when the excitation frequency shifts continuously from 20.4 Hz to 19.6 Hz. 
\subsubsection{Single-harmonic excitation tests}
\begin{figure}[h]
    \centering
    \includegraphics[width=.79\textwidth]{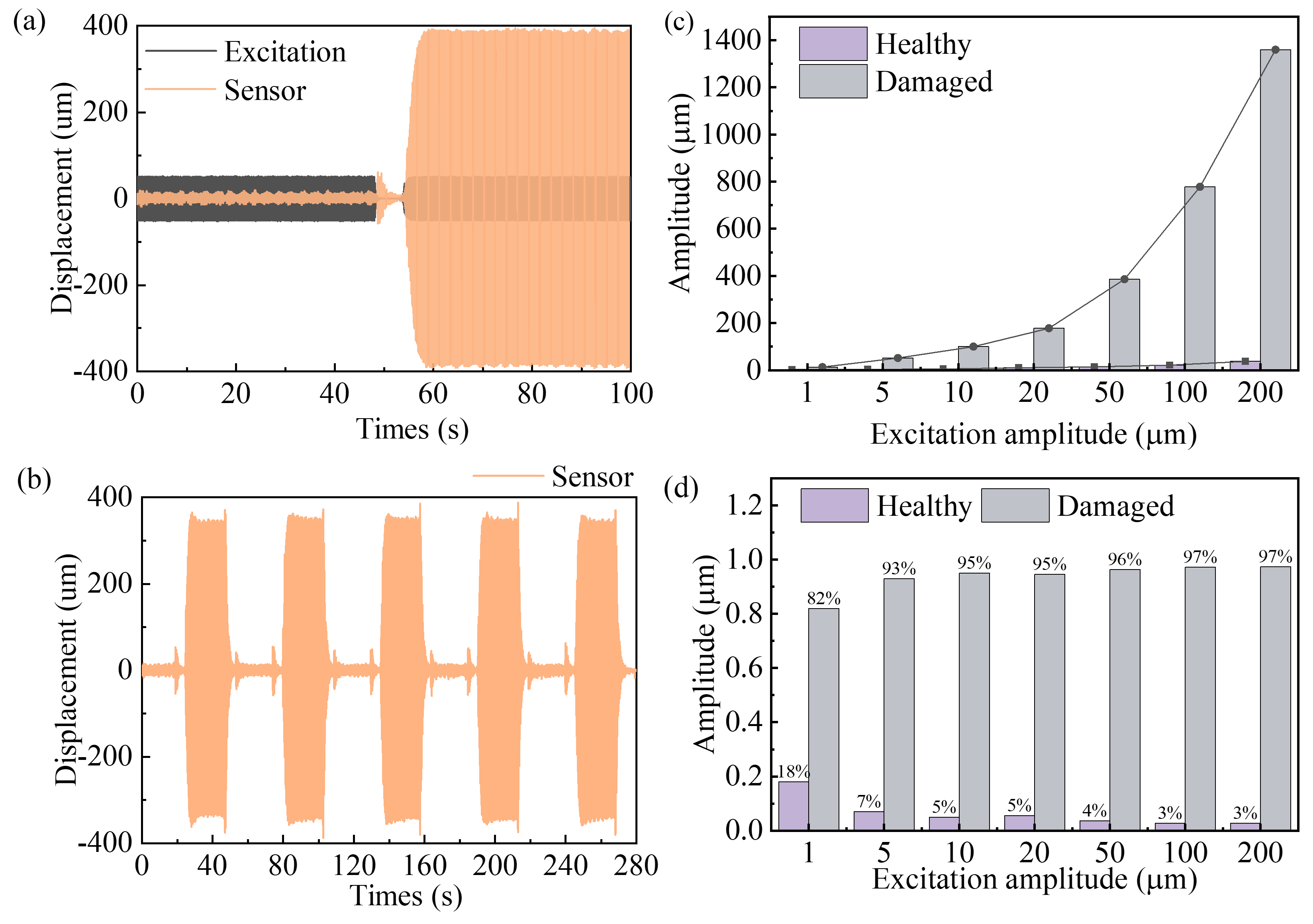}
    \caption{Single-frequency harmonic wave excitation experiment: (a) Emulated structural responses (excitation) and MM-sensor responses under healthy (0-50 seconds) and damaged (50-100 seconds) conditions; (b)A five "healthy-damaged" (or "0-1") cycles realized by the proposed MM-sensor (starting from time t = 0 as logic 0); (c) The sensor amplitudes in the 0 and 1 states under different excitation amplitudes; (d) The sensor normalized amplitudes in the 0 and 1 states under different excitation amplitudes.}
    \label{fig:exp222}
\end{figure}
We first consider an abrupt damage scenario, which is imitated by a sudden change of the harmonic frequency of the shaker. Specifically, a sinusoidal excitation with an amplitude of 0.1 mm (100 $\mu m$) is applied to the MM-sensor, with a total duration of 100 seconds. Between 0 and 50 seconds, the excitation frequency is constantly 20.4 Hz.  At the 50th second, the excitation frequency suddenly drops to 19.6 Hz, and continues for the remaining 50 seconds. The displacement of the probe point on the MM-sensor is shown in Figure \ref{fig:exp222}(a). It can be observed that when the excitation frequency is 20.4 Hz (imitating the healthy structure), the probe point displacement amplitude is approximately 34 $\mu m$. When the excitation frequency suddenly changes to 19.6 Hz (imitating the damaged structure), the MM-sensor responds promptly. The displacement amplitude increases to around 365 $\mu m$, approximately 3.65 times higher than the excitation amplitude. Moreover, this "healthy-damaged" characterization state is consistent across five cycles (Figure \ref{fig:exp222}(b)), which is similar to the performance of a NOT gate,  realized by a phonon transistor-like device based on magnetic coupling \cite{bilal2017bistable}. This indicates the potential for developing mechanical computing devices analogous to electronic systems (such as diodes and transistors).

To test the effective applicability range of the MM-sensor, 7 more excitation amplitudes (1, 5, 10, 20, 50, 100, and 200 $\mu m$) were tested. The displacement amplitudes of the probe point before and after structural damage are shown in Figure \ref{fig:exp222}(c), and the normalized metrics under healthy and damaged conditions are shown in Figure \ref{fig:exp222}(d). It is observed that as the excitation vibration amplitude increases, it becomes more evident to distinguish between the healthy and damaged states. For example, when the excitation amplitude is 5 $\mu m$, the metric is 93\%, whereas when the excitation amplitude is 200 $\mu m$, the metric is 97\%, having a higher accuracy.  It is noted that when the excitation amplitude is as small as 1 $\mu m$, the sensor's response amplitude in the healthy state reaches 2.92 $\mu m$ (> 1 $\mu m$), indicating a failure in binary classification, as the sensor is unable to effectively suppress the input excitation. Nonetheless, the proposed MM-sensor generally exhibits distinct vibration amplitudes before and after changes in the excitation frequency, demonstrating a working precision of up to 5 $\mu m$.

\subsubsection{Multi-harmonic excitation tests}

To evaluate the classification performance of the proposed MM-sensor under a more realistic structural response, multi-harmonic excitation tests were conducted. Two different structural responses were considered. The first type of structural response (healthy state) was generated by superimposing multiple sine waves of 20.4 Hz, 67.6 Hz, and 92.4 Hz, with corresponding displacement amplitudes of 0.0544 mm, 0.00610 mm, and 0.00345 mm. 
The three frequency components of the second structural response in a healthy state are 20.4 Hz, 99 Hz, and 153 Hz, with corresponding displacement amplitudes of 0.0353 mm, 0.000406 mm, and 0.000128 mm. For the above two scenarios, the damage to the structure is simulated by a reduction of approximately 3\% in all the three fundamental frequencies. 

\begin{figure}[h]
    \centering   \includegraphics[width=.89\columnwidth]{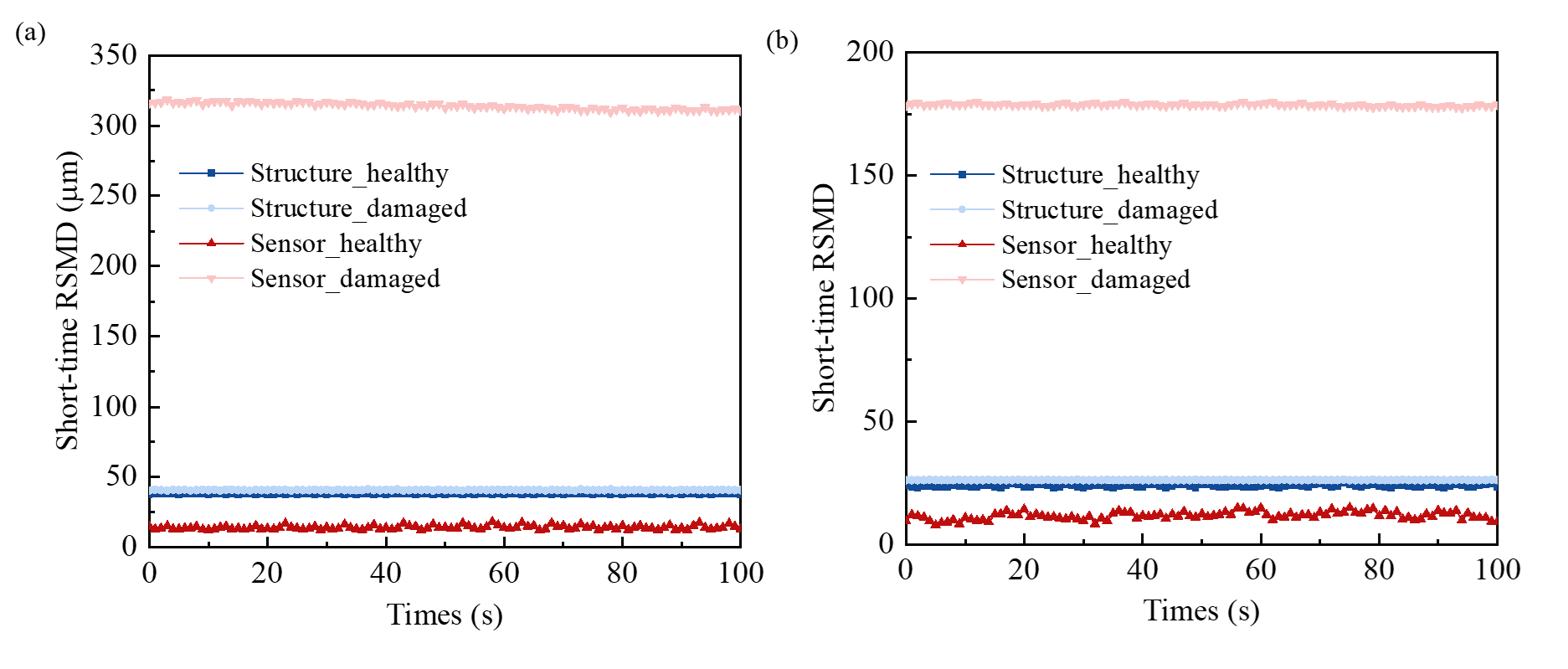}
    \caption{Short-time RMSD of the structure and MM-sensor under multi-harmonic excitation: (a) Result for multi-harmonic excitation 1 (case where high frequencies dominate); (b) Result for multi-harmonic excitation 2 (case where low frequencies dominate).}
    \label{fig:expmutisin}
\end{figure}

To perform the classification, we used the short-time root mean square displacement (RMSD in Eq.\eqref{eq:RMSD}, computed every 1 second), and the results are shown in Figure \ref{fig:expmutisin}. In this figure, the red and blue curves represent the sensor vibration and the structure vibration, respectively. The RMSD curves for the healthy and damaged states of the structure itself are almost indistinguishable. Evidently, the RMSD of the MM-sensor in a healthy state remains at a low level close to 0, while the RMSD in a damaged state is significantly higher than in a healthy state, demonstrating a sensitive ability to distinguish between the two conditions. Additionally, we observed mild fluctuations in the RMSD curve of the sensor in the healthy state. This is likely due to the probe point response remaining at the micron level, making it susceptible to fluctuations caused by environmental noise. 
\begin{figure}
    \centering
    \includegraphics[width=.99\textwidth]{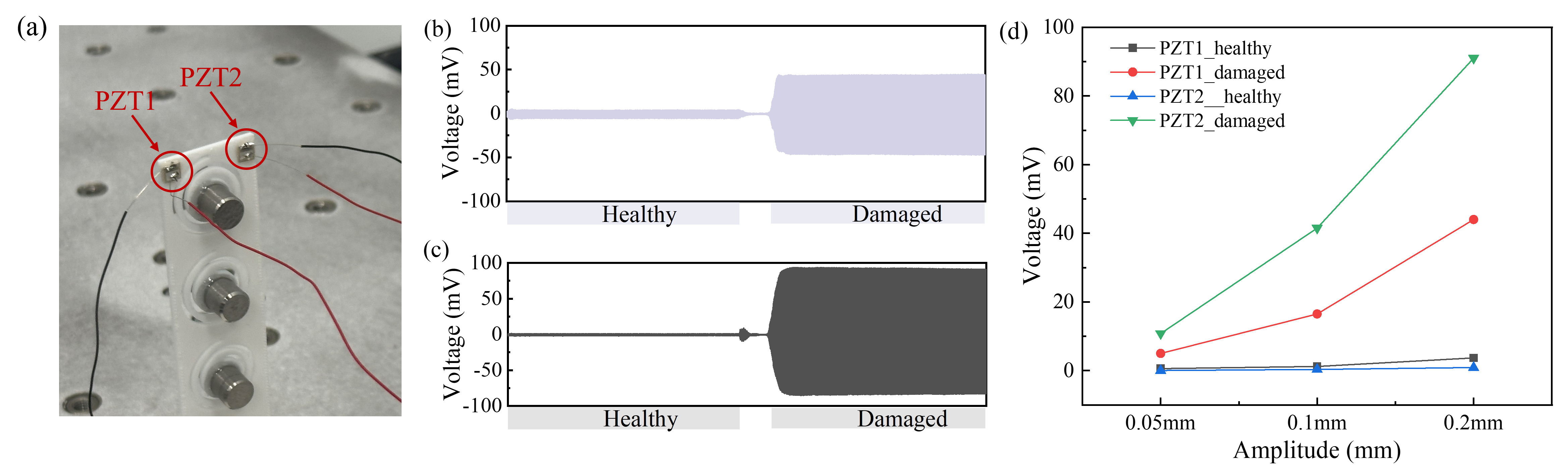}
    \caption{(a) Schematic diagram of piezoelectric element installation; (b) The voltage-time curve of the piezoelectric element 1 (PZT$1$), the first 50 seconds and the last 50 seconds represent the structure in healthy and damaged states, respectively; (c) The voltage-time curve of the piezoelectric element 2 (PZT$2$); (d) The voltage values under different magnitudes of excitation.}
    \label{fig:exp_pzt}
\end{figure}
Figure \ref{fig:exp_pzt}(a) shows the piezoelectric experimental setup, where two piezoelectric elements (PZT$1$ and PZT$2$) are glued on the left and right sides of the end of the plate. Figure \ref{fig:exp_pzt}(b) shows the real-time voltage curves of the two piezoelectric elements under healthy and damaged conditions (with an excitation amplitude of 0.2 mm). The figure clearly indicates that the voltage values are significantly amplified to higher values under the damaged condition.  It is worth noting that the voltage value of PZT$2$ is greater than that of PZT$1$, as the structure of the plate -- the spiral groove -- is not perfectly symmetric. This is consistent with the numerical analysis pattern in Figure \ref{fig:pzt12}(d). In terms of the voltage values, when the excitation amplitude is 0.2 mm, the maximum voltage reaches 91 mV, which does not meet the reference value of 2 V presented in Section \ref{pzt}. This is related to the introduction of low stress transfer efficiency and additional uneven stress distribution due to imperfect adhesion \cite{qing2006effect}. However, to achieve a sufficiently high voltage for triggering an alarm circuit in the event of structural damage, stacking multiple piezoelectric elements is a possible solution.

\section{Conclusions and Discussion}\label{conclusion}
We propose a MM-sensor for mechanical in-sensor computing, and the proof-of-concept is to physically process structural vibration signals, performing direct binary structural damage classification. The MM-sensor is based on a locally resonant metamaterial plate, which is composed of a plate and several metallic resonators. Structural damage-induced natural frequency shifts are captured by the metamaterial’s bandgap behavior, which functions as a mechanical classifier. By mapping mechanical responses to classification outcomes, the proposed MM-sensor achieves binary structural damage classification without requiring analog-to-digital converters (ADCs) or external power supplies, albeit the dynamic model of the proposed metamaterial is relatively simple. By adjusting the geometric parameters of the resonators, the MM-sensor can be programmable, working for engineering systems whose first natural frequency resides in the range of 9.54 - 81.86 Hz. Experimental results indicate that the manufacturing precision of these small-scale resonators is extremely high. In the future, active tuning mechanisms, such as piezoelectric actuators or magnetic field control, could be integrated to enable real-time dynamic adjustment of the MM-sensor's performance without requiring hardware (geometric) modifications, thereby addressing the challenges posed by varying monitoring environments. 


Although the current result is limited to preliminary binary classification tasks, this work has taken the first step towards structural health monitoring based on mechanical in-sensor computing, or physical neural networks that mimic digital neural networks. The binary classification focuses on the significant changes in vibration patterns triggered by structural stiffness damage; however, the sensor's sensitivity to early-stage subtle damage or environmental disturbances (e.g., temperature fluctuations) still needs optimization. For multiple-class damage classification (such as distinguishing crack types, locations, or severity), the number of resonators is expected to increase to provide more internal vibration modes that enhance the expressiveness of the MM-sensor. 
Finally, the efficiency of the energy conversion from mechanical energy to electrical energy is critical for the development of sustainable intelligent monitoring solutions. In the future, integrating high-efficiency energy conversion materials and low-power consumption circuits will be crucial for achieving truly self-powered and reliable damage alert systems.
\printcredits

\section*{Acknowledgments}
The authors wish to express their gratitude for the financial support received from Guangdong Provincial Fund - Special Innovation Project (2024KTSCX038);  Research Grants Council of Hong Kong through the Research Impact Fund (R5006-23); the Guangdong Provincial Key Lab of Integrated Communication, Sensing and Computation for Ubiquitous Internet of Things (No.2023B1212010007).


\newpage
\appendix
\renewcommand{\thesection}{Appendix \Alph{section}}
\renewcommand{\thefigure}{\Alph{section}.\arabic{figure}}
\section{Band structure evaluation}\label{app:a}
\setcounter{figure}{0} 
\begin{figure}
    \centering
    \includegraphics[width=.95\textwidth]{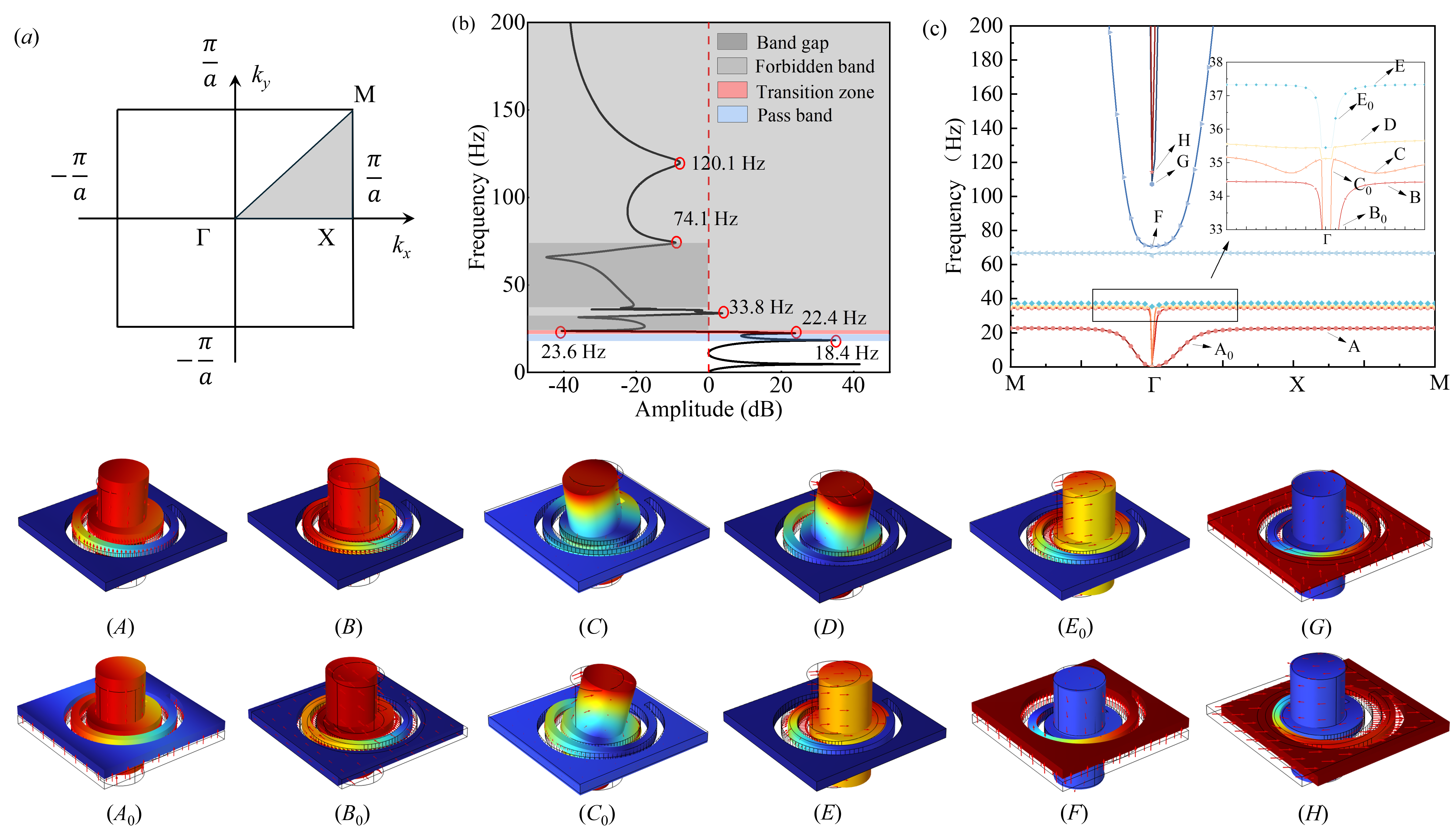}
    \caption{Analysis of Bandgap Mechanism and Vibration Characteristics: (a) First Brillouin zone with the irreducible Brillouin zone highlighted; (b) Transmittance curve of point on the center of the top surface of the fifth stainless steel cylinder;(c) Band structure; (A-H) Eigenmode shapes and displacement vector.}
    \label{fig:Bandstructure}
\end{figure}
In order to provide more insights into the dynamic behavior of the designed MM-sensor, a unit cell analysis was conducted. Wave propagation in periodic structures can be characterized by analyzing a representative unit cell subjected to periodic boundary conditions. According to the theory of elastic wave propagation, the governing equation for time-harmonic motion in a homogeneous and isotropic solid is given by the following equation \cite{kudela2023deep}:
\begin{equation}
(\lambda + 2\mu) \nabla(\nabla \cdot \mathbf{u}) - \mu \nabla \times \nabla + \rho \omega^2 \mathbf{u} = 0 
\end{equation}
where \( \lambda \) and \( \mu \) are the Lamé parameters, \( \rho \) represents the mass density, \( \mathbf{u} \) is the displacement vector, \( \omega \) denotes the angular frequency, and \( \nabla \) is the gradient operator. 
According to Bloch’s theorem, the displacement field corresponding to an eigenmode can be expressed as:
\begin{equation}
\mathbf{u}(\mathbf{r}) = e^{-i(\mathbf{k} \cdot \mathbf{R})} \mathbf{u}(\mathbf{r+R}) 
\end{equation}
where \( \mathbf{u}(\mathbf{r}) \) denotes the general displacement field, and \( \mathbf{k} = (k_x, k_y) \) is the Bloch wavenumber vector, and \(\mathbf{R}\) is the lattice constant vector. 
By substituting Eq.(8) into Eq.(7), we arrive at an eigenvalue problem formulated as:
\begin{equation}
\begin{bmatrix}
\mathbf{K} -\omega^2 \mathbf{M}
\end{bmatrix}
\mathbf{u} = 0 
\end{equation}

By taking values along the irreducible Brillouin zone boundary  \( ( M\rightarrow\Gamma\rightarrow X \rightarrow M) \) of the unit cell (as in Figure \ref{fig:Bandstructure}(a)) and then solving the eigenvalue problem, the resulting band structure diagram can intuitively reveal the frequency ranges corresponding to passbands and bandgaps. 
By applying Bloch-Floquet periodic boundary conditions to the unit cell boundaries, the bandgap structure calculated using the Structural Mechanics module in COMSOL Multiphysics is shown in Figure \ref{fig:Bandstructure}(c). 
In the \(  \Gamma \rightarrow X \) region, several nearly flat energy bands are clearly present, specifically the parts of the straight bands A-E that are farther from the \( \Gamma \) point. 
The group velocities are essentially zero for flat bands, which suggests that these modes do not carry energy across the structure.
Instead, the corresponding vibrational modes are strongly localized in space, often trapped within the unit cells (e.g., inside resonators).
In essence, the presence of flat bands signifies the occurrence of local resonances, often associated with the formation of bandgaps.
Figure \ref{fig:Bandstructure}(A-H) presents the calculated vibration modes at selected points on the band structure diagram. 
Mode $A$ corresponds to the transverse vibration mode, while modes $B$, $E$, and $E_0$ correspond to the longitudinal and lateral vibration modes within the plane of the plate. Furthermore, in Figure \ref{fig:Bandstructure}(b), the peak frequencies of the transmittance curve are 18.4 Hz, 33.8 Hz, 74.1 Hz, and 120.1 Hz, corresponding to modes $A_0$, $B_0$, $F$, and $G$, respectively. 
This explains the pronounced vibrations observed in the cylindrical mass and plates at the corresponding frequencies. 
For modes $C$, $C_0$, and $D$, the cylindrical mass undergoes out-of-plane torsional vibrations around the plane of the plate. 

In general, in these modes extracted from the flat bands, the plate exhibits negligible vibration, while the cylindrical mass undergoes both out-of-plane and in-plane motions. This indicates that the vibration energy is confined within the cylindrical mass, which is the well-known phenomenon referred to as "local resonance."


\section{Parametric study for programmability}\label{app:b}

\setcounter{figure}{0} 
\begin{figure}
    \centering
    \includegraphics[width=.7\textwidth]{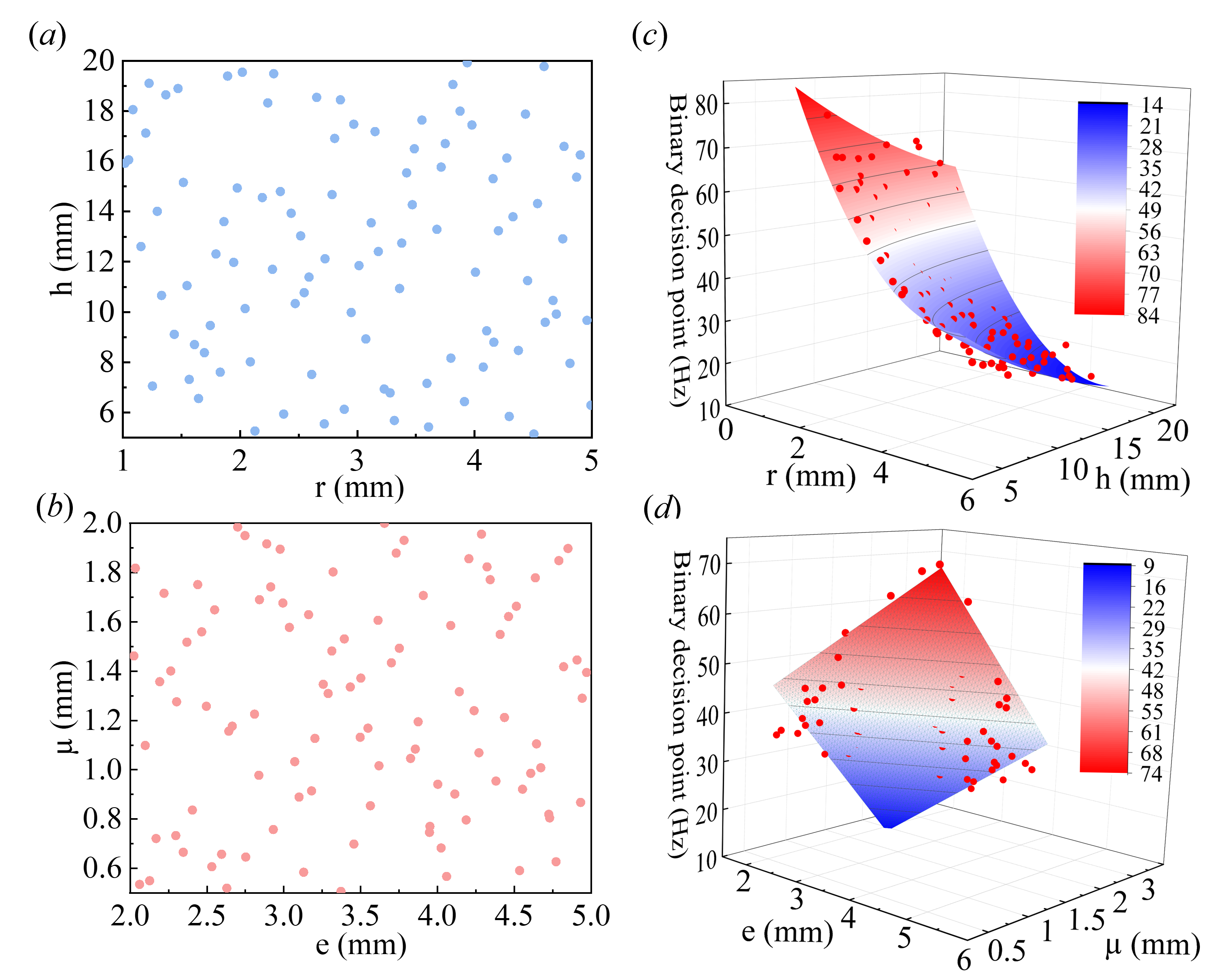}
    \caption{Parametric study for programmability: (a) Sampling distribution of the r and h; (b) Sampling distribution of the \( \mu\) and $e$ (c) The relationship between BDPs and the radius r and height h of the stainless steel cylinder; (d) The relationship between BDPs and the groove \( \mu\) and Board thickness $e$ of the PLA plate.}
    \label{fig:parametric}
\end{figure}
This section investigates the programmability of MM-sensors based on a parametric study of geometric parameters.
Based on the analysis of ~\ref{app:a}, it is known that the boundary values of the bandgap are determined by the frequency of the local resonance mode. By changing the geometric parameters of the steel cylinder and the plate, the frequency corresponding to the local resonance mode can be altered, thereby adjusting the upper and lower bounds of the bandgap, accommodating the MM-sensor for monitoring structures with distinct natural frequencies. In order to facilitate the design of the proposed MM-sensor, COMSOL simulations were used to investigate the effects of the resonator's dimensions on height $h$, width $r$, the thickness of the PLA plate $e$, and the width of the spiral groove $\mu$. Specifically, the transmittance of the MM-sensor with varying geometric parameters was calculated, and the frequency where the transmittance (in dB) becomes zero (marking the transition between the passband and attenuation band) was identified as the corresponding binary decision point (BDP) value for the classification task.
To prepare the dataset, the first set of system parameters ($r$, $h$) and the second set ($e$, $\mu$) were generated separately using the Latin Hypercube Sampling method. Figures \ref{fig:parametric}(a-b) visualize the distribution of the generated 100 samples.
Figures \ref{fig:parametric}(c) and (d) show the relationship between BDP and the geometric parameters. The red scatter points in these figures represent the samples in the dataset, while the smooth surface obtained through Gaussian fitting reflects the overall trend of these samples. From Figure \ref{fig:parametric}(c), it can be observed that the BDP value exhibits a distinct nonlinear distribution trend with changes in $r$ and $h$, with a total variation ranging from 15.65 Hz to 77.66 Hz. Specifically, as $r$ and $h$ increase, the BDP value shows a decreasing trend; the slope of the surface corresponding to $r$ is significantly greater than that of $h$, indicating that the BDP value is more sensitive to $r$, especially when $r$ approaches its minimum value. However, the BDP value shows a relatively linear decrease with the increase of e and the decrease of $m$, spanning a range from approximately 18.62 Hz to 73.99 Hz. Furthermore, the range of the BDP values varies from 9.54 Hz to 81.86 Hz while these four geometric parameters is changed simultaneously. Based on the above findings, it is not difficult to imagine that the simultaneous variation of these four geometric parameters has a higher-dimensional non-linear comprehensive effect on BDP, thereby increasing the complexity of subsequent optimization and design.

\section{Training processes of forward and inverse models}\label{app:c}
\setcounter{figure}{0} 

\begin{algorithm}[!h]
\caption{Training Process for Forward Modeling}
\label{alg:forward_training}
\begin{algorithmic}[1]
\State \textbf{Inputs}: Training dataset \( \mathcal{D}_{\text{train}} = \{(\mathbf{g}_i, y_i)\}_{i=1}^{n_t} \), forward model \( f_{\theta} \), epochs \( N \), batch size \( B \)
\For{\texttt{epoch} = 1 to \( N \)}
    \For{\texttt{batch} of size \( B \) in \( \mathcal{D}_{\text{train}} \)}
        \State Predict: \(\hat{{y}} = f_{\theta}(\mathbf{g})\)
        \State Compute MSE loss: \( \mathcal{L}_{\text{MSE}} = \frac{1}{B} \sum_{i=1}^B (\hat{{y}}_i - {y}_i)^2 \)
        \State Add $\ell_2$ regularization: \( \mathcal{L} = \mathcal{L}_{\text{MSE}} + \lambda \sum_{\text{residual params}} \|\theta\|_2^2 \), with \(\lambda = 10^{-3}\)
        \State Update \( f_{\theta}\) via gradient descent w.r.t. \(\nabla \mathcal{L}\)
    \EndFor
    \State Evaluate MSE loss \( \mathcal{L}_{\text{val}} \) on the validation dataset
    \State Update learning rate via scheduler based on \( \mathcal{L}_{\text{test}} \)
    \If{\( \mathcal{L}_{\text{val}} \) is lowest}
        \State Save model state as \(f_{\theta^*}\)
    \EndIf
\EndFor
\State \textbf{Output}: Trained model \(f_{\theta^*}\)
\end{algorithmic}
\end{algorithm}

\begin{algorithm}[!h]
\caption{Training Process for Inverse Modeling}
\label{alg:inverse_training}
\begin{algorithmic}[1]
\State \textbf{Inputs}: Target feature point \({y}_{\text{target}}\), trained forward model \(f_{\theta^*}\), 
rials \( T \), iterations \( N \)
\For{\texttt{trial} = 1 to \( T \)}
    \State Randomly initialize geometric parameters: \(\mathbf{g} \sim \mathcal{N}(0, 1),\) and \(\mathbf{g}  \in \mathbb{R}^4 \)
    \For{\texttt{iteration} = 1 to \( N \)}
        \State Predict: \(\hat{{y}} = f_{\theta^*}(\mathbf{g})\)
        \State Compute MSE loss: \( \mathcal{L} = (\hat{{y}} - {y}_{\text{target}})^2 \)
        \State Update \(\mathbf{g}\) via gradient descent w.r.t. \(\nabla \mathcal{L}\)
        \State Clamp \(\mathbf{g}\) within scaled bounds of \((\mathbf{g}_{\text{min}}, \mathbf{g}_{\text{max}})\)
    \EndFor
    \If{\( \mathcal{L} < \mathcal{L}_{\text{best}} \)}
        \State Update \( \mathcal{L}_{\text{best}} = \mathcal{L} \), \(\mathbf{g}_{\text{best}} = \mathbf{g}\)
    \EndIf
\EndFor
\State \textbf{Output}: Predicted geometric parameters \(\mathbf{g}_{\text{best}}\)
\end{algorithmic}
\end{algorithm}

This appendix provides a detailed description of the deep learning model used for forward and inverse MM-sensors.
The forward model aims to predict the binary decision point (BDP) \(y\) given the geometric parameters \(\mathbf{g}\). The training dataset consists of pairs \((\mathbf{g}_i, y_i)\), where \(\mathbf{g}_i \in \mathbb{R}^4\) and \(y_i \in \mathbb{R}^1\). The data is preprocessed using standard scaling, and Gaussian noise (mean 0, standard deviation 0.01) is added to the training inputs for augmentation. The model is trained using the mean squared error (MSE) loss with $\ell_2$ regularization applied to the residual blocks' parameters. The Adam optimizer with a learning rate of 0.001 and a ReduceLROnPlateau scheduler (factor 0.5, patience 50) are used to optimize the model over 5000 epochs. The best model state, based on the lowest test loss, is saved for inverse design. The training process is formalized in Algorithm~\ref{alg:forward_training}.

Inverse modeling seeks to predict the geometric parameters \(\mathbf{g}\) given a target binary decision point \(y_{\text{target}}\). The trained forward model \(f_{\theta^*}\) is implemented as the surrogate model of forward simulation.
The process involves gradient-based optimization over multiple trials (5 trials, each with 2000 iterations) to minimize the MSE between the predicted feature point \(f_{\theta}(\mathbf{g})\) and the target \(y_{\text{target}}\). Randomly initialized geometric parameters are optimized using the Adam optimizer with a learning rate of 0.001, constrained within the scaled bounds of the training data. The best trial, based on the lowest loss, provides the predicted geometric parameters, which are then inverse-transformed to the original scale. The inverse prediction process is formalized in Algorithm~\ref{alg:inverse_training}.

\begin{figure}[h]
    \centering
    \includegraphics[width=.99\textwidth]{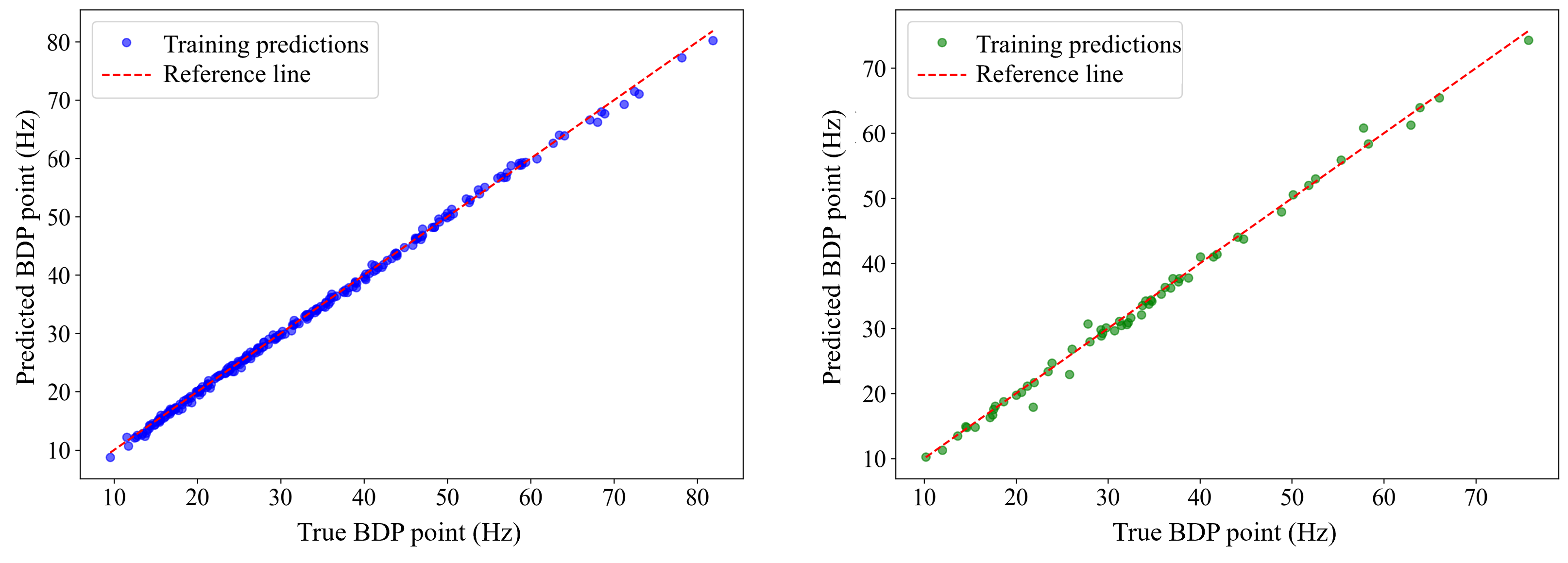}
    \caption{Result of forward model: (a) Training set; (b) Test set.}
    \label{fig:forwardresult}
\end{figure}

As for the design task for the Section \ref{result}. Firstly, a dataset composed of a training set of 244 samples (80\%) and a testing set of 61 samples (20\%) is used to train the forward model. The scatter plots in Figure \ref{fig:forwardresult} compare the predicted feature point values against their true values for the training and testing sets, respectively. The root mean square errors (RMSE) of the training set and test set are 0.0338 and 0.0705. The scatter plot of training stage shows a tight clustering of points around the ideal line (red dashed line), indicating high predictive accuracy and robust fitting to the training data. The scatter plot for the testing set demonstrates a strong linear relationship between true and predicted values, albeit with slightly greater dispersion compared to the training set, which is practically acceptable.

\begin{figure}[h]
    \centering
    \includegraphics[width=.99\textwidth]{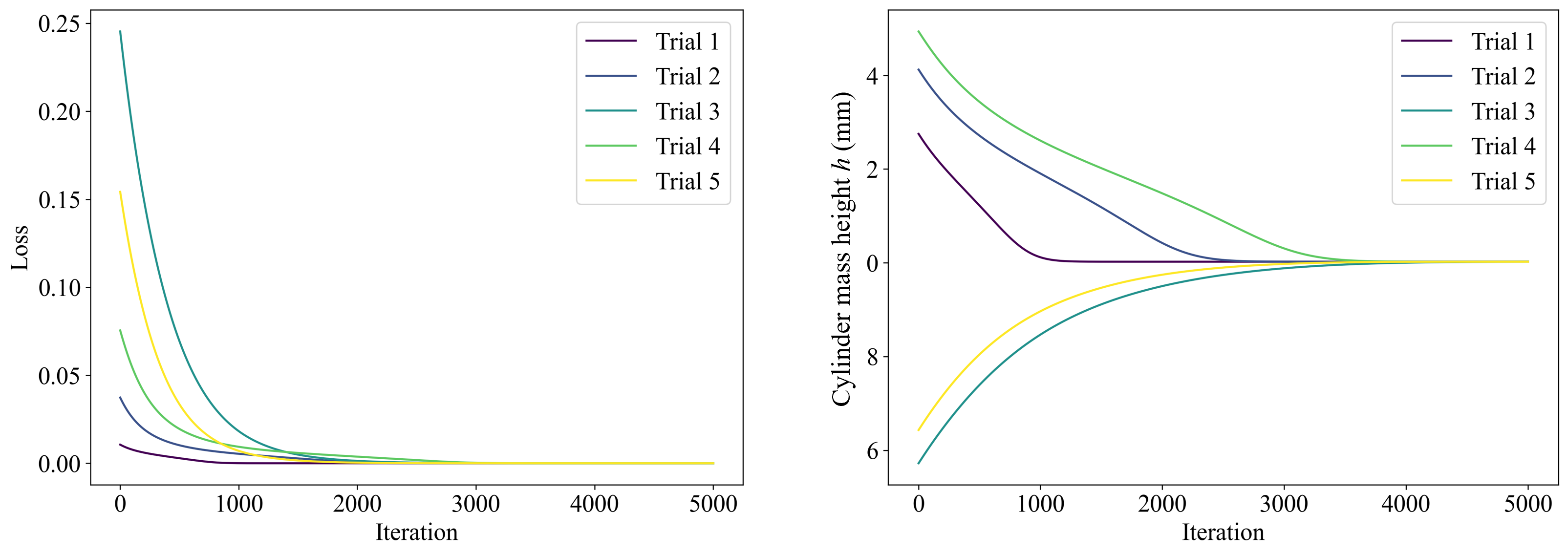}
    \caption{Result of inverse model: (a) Loss convergence curve for 5 Trials; (b) Trajectories of predicted value for 5 Trials.}
    \label{fig:inverseresult}
\end{figure}

In inverse design task, three geometric parameters 
($e$ = 2 mm, $\mu$ = 1.5 mm, $r$ = 5 mm) remain unchanged, only the fourth parameter $h$ needs to be optimized. To illustrate the robustness of the inverse design optimization, the loss curve and trajectories of the steel cylinder height $h$ across 5 trials are plotted against optimization iterations. As in Figure \ref{fig:inverseresult}, each trial begins with a randomly initialized parameter value, and the convergence of all trajectories to a nearly identical final value (10.023 mm) with final loss below 0.0001, demonstrating the stability of the inverse model.

\newpage
\bibliographystyle{unsrt}


\begin{thebibliography}{10}

\bibitem{li2024mechanics}
Xuyang Li, Hamed Bolandi, Mahdi Masmoudi, Talal Salem, Ankush Jha, Nizar Lajnef, and Vishnu~Naresh Boddeti.
\newblock Mechanics-informed autoencoder enables automated detection and localization of unforeseen structural damage.
\newblock {\em Nature Communications}, 15(1):9229, 2024.

\bibitem{huang2024tunable}
Rongjuan Huang, Yunfei He, Juan Wang, Jindou Zou, Hailan Wang, Haodong Sun, Yuxin Xiao, Dexin Zheng, Jiani Ma, Tao Yu, et~al.
\newblock Tunable afterglow for mechanical self-monitoring 3d printing structures.
\newblock {\em Nature Communications}, 15(1):1596, 2024.

\bibitem{ri2024drone}
Shien Ri, Jiaxing Ye, Nobuyuki Toyama, and Norihiko Ogura.
\newblock Drone-based displacement measurement of infrastructures utilizing phase information.
\newblock {\em Nature Communications}, 15(1):395, 2024.

\bibitem{rosales2017data}
Mary~J Rosales and Ranjith Liyanapathirana.
\newblock Data driven innovations in structural health monitoring.
\newblock In {\em Journal of Physics: Conference Series}, volume 842, page 012012. IOP Publishing, 2017.

\bibitem{xie2025smart}
Guihua Xie, Shu-yang WANG, Hong-yun XIA, and Shi-quan LI.
\newblock Smart-monitoring capabilities of cfrp strips in civil structures based on width-to-length ratio optimization.
\newblock {\em Smart Materials and Structures}, 2025.

\bibitem{zhou2024structural}
Mingyuan Zhou and Zhilu Lai.
\newblock Structural damage classification under varying environmental conditions and unknown classes via open set domain adaptation.
\newblock {\em Mechanical Systems and Signal Processing}, 218:111561, 2024.

\bibitem{fan2020structural}
Zhichun Fan, Xingzhong Diao, Kangjia Hu, Yong Zhang, Zhiyong Huang, Yanbo Kang, and He~Yan.
\newblock Structural health monitoring of metal-to-glass--ceramics penetration during thermal cycling aging using femto-laser inscribed fbg sensors.
\newblock {\em Scientific Reports}, 10(1):12330, 2020.

\bibitem{qu2024high}
Zhan Qu, Zhenjun Zhang, Rui Liu, Ling Xu, Yining Zhang, Xiaotao Li, Zhenkai Zhao, Qiqiang Duan, Shaogang Wang, Shujun Li, et~al.
\newblock High fatigue resistance in a titanium alloy via near-void-free 3d printing.
\newblock {\em Nature}, 626(8001):999--1004, 2024.

\bibitem{cronin2024bridging}
Liam Cronin, Soheil Sadeghi~Eshkevari, Thomas~J Matarazzo, Sebastiano Milardo, Iman Dabbaghchian, Paolo Santi, Shamim~N Pakzad, and Carlo Ratti.
\newblock Bridging the gap: commodifying infrastructure spatial dynamics with crowdsourced smartphone data.
\newblock {\em Communications Engineering}, 3(1):93, 2024.

\bibitem{bhattacharya2025optimal}
Ashmita Bhattacharya, Konstantinos~G Papakonstantinou, Gordon~P Warn, Lauren McPhillips, Melissa~M Bilec, Chris~E Forest, Rahaf Hasan, and Digant Chavda.
\newblock Optimal life-cycle adaptation of coastal infrastructure under climate change.
\newblock {\em Nature Communications}, 16(1):1076, 2025.

\bibitem{malekloo2022machine}
Arman Malekloo, Ekin Ozer, Mohammad AlHamaydeh, and Mark Girolami.
\newblock Machine learning and structural health monitoring overview with emerging technology and high-dimensional data source highlights.
\newblock {\em Structural Health Monitoring}, 21(4):1906--1955, 2022.

\bibitem{lee2017toward}
Eun~Kwang Lee, Moo~Yeol Lee, Cheol~Hee Park, Hae~Rang Lee, and Joon~Hak Oh.
\newblock Toward environmentally robust organic electronics: approaches and applications.
\newblock {\em Advanced Materials}, 29(44):1703638, 2017.

\bibitem{shagrir2022nature}
Oron Shagrir.
\newblock {\em The nature of physical computation}.
\newblock Oxford University Press, 2022.

\bibitem{von1993first}
John Von~Neumann.
\newblock First draft of a report on the edvac.
\newblock {\em IEEE Annals of the History of Computing}, 15(4):27--75, 1993.

\bibitem{mcmahon2023physics}
Peter~L McMahon.
\newblock The physics of optical computing.
\newblock {\em Nature Reviews Physics}, 5(12):717--734, 2023.

\bibitem{pfeffer2022hybrid}
Philipp Pfeffer, Florian Heyder, and J{\"o}rg Schumacher.
\newblock Hybrid quantum-classical reservoir computing of thermal convection flow.
\newblock {\em Physical Review Research}, 4(3):033176, 2022.

\bibitem{yasuda2021mechanical}
Hiromi Yasuda, Philip~R Buskohl, Andrew Gillman, Todd~D Murphey, Susan Stepney, Richard~A Vaia, and Jordan~R Raney.
\newblock Mechanical computing.
\newblock {\em Nature}, 598(7879):39--48, 2021.

\bibitem{jeong2024cryogenic}
Jaeyong Jeong, Seong~Kwang Kim, Yoon-Je Suh, Jisung Lee, Joonyoung Choi, Joon~Pyo Kim, Bong~Ho Kim, Juhyuk Park, Joonsup Shim, Nahyun Rheem, et~al.
\newblock Cryogenic iii-v and nb electronics integrated on silicon for large-scale quantum computing platforms.
\newblock {\em Nature Communications}, 15(1):10809, 2024.

\bibitem{wright2022deep}
Logan~G Wright, Tatsuhiro Onodera, Martin~M Stein, Tianyu Wang, Darren~T Schachter, Zoey Hu, and Peter~L McMahon.
\newblock Deep physical neural networks trained with backpropagation.
\newblock {\em Nature}, 601(7894):549--555, 2022.

\bibitem{romera2018vowel}
Miguel Romera, Philippe Talatchian, Sumito Tsunegi, Flavio Abreu~Araujo, Vincent Cros, Paolo Bortolotti, Juan Trastoy, Kay Yakushiji, Akio Fukushima, Hitoshi Kubota, et~al.
\newblock Vowel recognition with four coupled spin-torque nano-oscillators.
\newblock {\em Nature}, 563(7730):230--234, 2018.

\bibitem{mead1990neuromorphic}
Carver Mead.
\newblock Neuromorphic electronic systems.
\newblock {\em Proceedings of the IEEE}, 78(10):1629--1636, 1990.

\bibitem{song2019additively}
Yuanping Song, Robert~M Panas, Samira Chizari, Lucas~A Shaw, Julie~A Jackson, Jonathan~B Hopkins, and Andrew~J Pascall.
\newblock Additively manufacturable micro-mechanical logic gates.
\newblock {\em Nature communications}, 10(1):882, 2019.

\bibitem{treml2018origami}
Benjamin Treml, Andrew Gillman, Philip Buskohl, and Richard Vaia.
\newblock Origami mechanologic.
\newblock {\em Proceedings of the National Academy of Sciences}, 115(27):6916--6921, 2018.

\bibitem{venstra2010mechanical}
Warner~J Venstra, Hidde~JR Westra, and Herre~SJ van~der Zant.
\newblock Mechanical stiffening, bistability, and bit operations in a microcantilever.
\newblock {\em Applied physics letters}, 97(19), 2010.

\bibitem{bilal2017bistable}
Osama~R Bilal, Andr{\'e} Foehr, and Chiara Daraio.
\newblock Bistable metamaterial for switching and cascading elastic vibrations.
\newblock {\em Proceedings of the National Academy of Sciences}, 114(18):4603--4606, 2017.

\bibitem{li2014granular}
Feng Li, Paul Anzel, Jinkyu Yang, Panayotis~G Kevrekidis, and Chiara Daraio.
\newblock Granular acoustic switches and logic elements.
\newblock {\em Nature communications}, 5(1):5311, 2014.

\bibitem{wu2024mechanical}
Lingling Wu, Yuyang Lu, Penghui Li, Yong Wang, Jiacheng Xue, Xiaoyong Tian, Shenhao Ge, Xiaowen Li, Zirui Zhai, Junqiang Lu, et~al.
\newblock Mechanical metamaterials for handwritten digits recognition.
\newblock {\em Advanced Science}, 11(10):2308137, 2024.

\bibitem{hughes2019wave}
Tyler~W Hughes, Ian~AD Williamson, Momchil Minkov, and Shanhui Fan.
\newblock Wave physics as an analog recurrent neural network.
\newblock {\em Science advances}, 5(12):eaay6946, 2019.

\bibitem{jiang2023metamaterial}
Tianxi Jiang, Tianqi Li, Hao Huang, Zhi-Ke Peng, and Qingbo He.
\newblock Metamaterial-based analog recurrent neural network toward machine intelligence.
\newblock {\em Physical Review Applied}, 19(6):064065, 2023.

\bibitem{hebb2005organization}
Donald~Olding Hebb.
\newblock {\em The organization of behavior: A neuropsychological theory}.
\newblock Psychology press, 2005.

\bibitem{momeni2024training}
Ali Momeni, Babak Rahmani, Benjamin Scellier, Logan~G Wright, Peter~L McMahon, Clara~C Wanjura, Yuhang Li, Anas Skalli, Natalia~G Berloff, Tatsuhiro Onodera, et~al.
\newblock Training of physical neural networks.
\newblock {\em arXiv preprint arXiv:2406.03372}, 2024.

\bibitem{farrar2007introduction}
Charles~R Farrar and Keith Worden.
\newblock An introduction to structural health monitoring.
\newblock {\em Philosophical Transactions of the Royal Society A: Mathematical, Physical and Engineering Sciences}, 365(1851):303--315, 2007.

\bibitem{zhang2013low}
Siwen Zhang, Jiu Hui~Wu, and Zhiping Hu.
\newblock Low-frequency locally resonant band-gaps in phononic crystal plates with periodic spiral resonators.
\newblock {\em Journal of Applied Physics}, 113(16), 2013.

\bibitem{zheng2023deep}
Xiaoyang Zheng, Xubo Zhang, Ta-Te Chen, and Ikumu Watanabe.
\newblock Deep learning in mechanical metamaterials: from prediction and generation to inverse design.
\newblock {\em Advanced Materials}, 35(45):2302530, 2023.

\bibitem{ia2016deep}
Goodfellow Ia.
\newblock Deep learning/ian goodfellow, yoshua bengio and aaron courville, 2016.

\bibitem{hutcheon2000resonance}
Bruce Hutcheon and Yosef Yarom.
\newblock Resonance, oscillation and the intrinsic frequency preferences of neurons.
\newblock {\em Trends in neurosciences}, 23(5):216--222, 2000.

\bibitem{senan2007brief}
Nur Adila~Faruk Senan.
\newblock A brief introduction to using ode45 in matlab.
\newblock {\em University of California at Berkeley, USA}, 2007.

\bibitem{lai2016moving}
Zhilu Lai, Ying Lei, Songye Zhu, You-Lin Xu, Xiao-Hua Zhang, and Sridhar Krishnaswamy.
\newblock Moving-window extended kalman filter for structural damage detection with unknown process and measurement noises.
\newblock {\em Measurement}, 88:428--440, 2016.

\bibitem{zhao20223d}
Pengcheng Zhao, Kai Zhang, Liyuan Qi, and Zichen Deng.
\newblock 3d chiral mechanical metamaterial for tailored band gap and manipulation of vibration isolation.
\newblock {\em Mechanical Systems and Signal Processing}, 180:109430, 2022.

\bibitem{qing2006effect}
Xinlin~P Qing, Hian-Leng Chan, Shawn~J Beard, Teng~K Ooi, and Stephen~A Marotta.
\newblock Effect of adhesive on the performance of piezoelectric elements used to monitor structural health.
\newblock {\em International Journal of Adhesion and Adhesives}, 26(8):622--628, 2006.

\bibitem{kudela2023deep}
Pawe{\l} Kudela, Abdalraheem Ijjeh, Maciej Radzienski, Marco Miniaci, Nicola Pugno, and Wieslaw Ostachowicz.
\newblock Deep learning aided topology optimization of phononic crystals.
\newblock {\em Mechanical Systems and Signal Processing}, 200:110636, 2023.

\end{thebibliography}
\bibliographystyle{unsrt} 


\end{document}